\documentclass[11pt]{article}

\usepackage{acl}

\usepackage{times}
\usepackage{latexsym}
\usepackage{algorithm}
\usepackage{algpseudocode}
\usepackage{amsmath} 
\usepackage[T1]{fontenc}

\usepackage[utf8]{inputenc}

\usepackage{microtype}

\usepackage{inconsolata}
\usepackage{booktabs}
\usepackage{booktabs}  
\usepackage{tabularx}  
\usepackage{ragged2e}  
\usepackage{graphicx}
\usepackage{multicol}
\usepackage{enumitem}

\title{SWE-Hub: A Unified Production System for Scalable, Executable Software Engineering Tasks}

\author{
\textbf{Qianfan Team, Baidu Inc.}
}

\begin{document}
\maketitle
\begin{abstract}
Progress in software-engineering agents is increasingly constrained by the scarcity of executable, scalable, and realistic data for training and evaluation. This scarcity stems from three fundamental challenges in existing pipelines: environments are brittle and difficult to reproduce across languages; synthesizing realistic, system-level bugs at scale is computationally expensive; and existing data predominantly consists of short-horizon repairs, failing to capture long-horizon competencies like architectural consistency. We introduce \textbf{SWE-Hub}, an end-to-end system that operationalizes the data factory abstraction by unifying environment automation, scalable synthesis, and diverse task generation into a coherent production stack. At its foundation, the \textbf{Env Agent} establishes a shared execution substrate by automatically converting raw repository snapshots into reproducible, multi-language container environments with standardized interfaces. Built upon this substrate, \textbf{SWE-Scale} engine addresses the need for high-throughput generation, combining cross-language code analysis with cluster-scale validation to synthesize massive volumes of localized bug-fix instances. \textbf{Bug Agent} generates high-fidelity repair tasks by synthesizing system-level regressions involving cross-module dependencies, paired with user-like issue reports that describe observable symptoms rather than root causes. Finally, \textbf{SWE-Architect} expands the task scope from repair to creation by translating natural-language requirements into repository-scale build-a-repo tasks. By integrating these components, SWE-Hub establishes a unified production pipeline capable of continuously delivering executable tasks across the entire software engineering lifecycle.

\end{abstract}

\section{Introduction}

Large language model (LLM) agents have become increasingly capable at writing code and repairing bugs, yet their progress on \textbf{repository-level} software engineering remains constrained by the scarcity of high-quality data~\citep{swebench, swebenchverified, swebenchpp, swebenchpro, skyworkswe, agentdataprotocol, swelego, swe-universe}. This scarcity arises because software engineering correctness is \textbf{execution-grounded}: determining whether a change is correct depends on the complex interplay of build systems, dependency graphs, runtime behavior, and test harnesses, rather than textual similarity or static diffs. Consequently, the prohibitive cost and inherent complexity of generating such execution-grounded data constitute a fundamental bottleneck, hindering the scalability of agent training and evaluation. Therefore, effective datasets must not only be large but also \textbf{executable by construction}, ensuring each instance is paired with a reproducible environment and a verifiable oracle. Crucially, effective data must also excel across three dimensions: cross-language reproducibility, system-level fault realism, and long-horizon task complexity.

Generating such comprehensive data is difficult because these three dimensions are misaligned in current pipelines: \textbf{First}, ensuring reproducibility across languages is a structural challenge~\citep{multiswebenche, qwen3codernext, swe-universe}. Real-world repositories are heterogeneous in their languages, toolchains, and test ecosystems; indeed, reproducing a single project often requires expert, manual environment debugging. \textbf{Second}, even within reproducible environments, current pipelines lack scalable, cross-language bug synthesis methods~\citep{swesmith, swesynth, swe-flow}. Consequently, much of today’s synthetic bug data remains local: perturbations are typically confined to a single function or file, which under-represents failures that arise from cross-module contracts, configuration interactions, or subtle regressions whose symptoms surface far from the root cause. \textbf{Third}, beyond fault complexity, the task scope remains limited. The dominant benchmark format focuses on short-horizon repair—given a failing signal, produce a patch—while under-measuring long-horizon competencies such as planning, dependency management, multi-file coherence, and repository-scale consistency~\citep{sweevo, swe-refactor, swe-perf}.

This paper argues that the right abstraction is not the manual data curation typical of existing benchmarks, but a \textbf{data factory}: a reproducible production system that continuously converts raw repositories into diverse, executable tasks. A practical data factory must satisfy three core requirements simultaneously: \textbf{(i) environment reliability}, which ensures the automated provisioning of deterministic, execution-ready environments for heterogeneous repositories~\citep{swe-universe, swe-rm}; \textbf{(ii) scalable data synthesis}, which enables the massive-scale generation  of data candidates by integrating automated synthesis algorithms with cluster-scale computing for parallel validation~\citep{swesynth, swe-flow, swelego}; and \textbf{(iii) task versatility}, which allows the factory to generate a wide spectrum of task types that range from localized repairs to repository construction~\citep{nl2repobench, katcoder}.

To operationalize the data factory abstraction, we introduce SWE-Hub, an end-to-end system that unifies the three core requirements through a shared execution substrate. At its foundation, the \textbf{Env Agent} automatically converts raw repository snapshots into reproducible, execution-ready environments. By standardizing container images and verification interfaces, this Agent transforms environment construction from a manual, ad-hoc prerequisite into a reusable asset.

Founded on this substrate, SWE-Hub implements three product lines that collectively fulfill the remaining requirements:

\begin{enumerate}
    \item \textbf{SWE-Scale: scalable synthesis for repair.} Building on SWE-Smith~\citep{swesmith}, SWE-Scale engine addresses the need for high-throughput generation by producing large volumes of localized bug-fix instances. It combines cross-language code analysis with automated bug injection and cluster-scale validation to efficiently synthesize and filter candidates based on execution signals.
    \item \textbf{Bug Agent: realistic repair task synthesis.} Bug Agent generates high-fidelity repair tasks by synthesizing system-level regressions that reflect cross-module dependencies. Unlike localized bug injection, it produces user-like issue reports focusing on observable symptoms rather than root causes, ensuring the repair tasks mirror real-world debugging scenarios.
    \item \textbf{SWE-Architect: versatile long-horizon tasks.} Extending the NL2Repo benchmark~\citep{nl2repobench}, SWE-Architect translates natural-language requirements into repository-scale construction tasks. This enables the factory to generate diverse training data that covers a wide spectrum of competencies—from localized fixes to global architectural planning—within a unified production pipeline.
\end{enumerate}

SWE-Hub consolidates previously fragmented efforts, which include environment automation, scalable synthesis, realism-driven bug generation, and long-horizon task design, into a coherent production stack with standardized interfaces, unified verification, and shared extensibility mechanisms. Rather than releasing a static dataset, we provide a production system capable of continuously generating executable tasks, scaling with compute, and adapting to new languages and task families.

We make the following contributions:
\begin{itemize}
\item \textbf{A unified data factory architecture.} We propose a system that standardizes the production of execution-grounded software engineering tasks. By integrating environment setup, candidate generation, and verification, this architecture enables compute-scalable data creation that is reproducible by construction.
\item \textbf{An automated reproducibility substrate.} We develop the Env Agent to act as a universal adapter for raw repositories. It autonomously provisions and verifies containerized environments, decoupling task generation from the complexities of language-specific toolchains and build systems.
\item \textbf{Scalable and realistic task synthesis mechanisms.} We introduce \textbf{SWE-Scale} engine to achieve high-throughput generation via parallelized validation, and \textbf{Bug Agent} to ensure high fidelity by creating system-level faults with decoupled symptoms and root causes.
\item \textbf{Expansion of task horizons.} We introduce \textbf{SWE-Architect} to generate "build-from-scratch" tasks at scale, producing training data that targets long-context competencies—such as architectural planning and global consistency—which are underrepresented in repair-centric datasets.
\end{itemize}

\section{Related Work}
\label{sec:related_work}

SWE-Hub is positioned at the intersection of (i) \textbf{execution-grounded} software engineering (SWE) benchmarks that rely on containerized evaluation, and (ii) \textbf{scalable task generation} methods that synthesize verifiable bug-fix instances over real repositories. In contrast to one-off benchmark releases, our work presents a \textbf{production system} designed to continuously transform raw repositories into executable task records through a shared execution substrate and multiple, specialized task product lines.

\subsection{Execution Substrates and Reproducible Evaluation Harnesses}
A significant body of recent literature evaluates repository-level agents using containerized harnesses that standardize dependency installation, test invocation, and outcome collection. SWE-bench~\citep{swebench} pioneered the issue-resolution setting over real GitHub repositories, providing a Docker-based harness for reproducible evaluation. Subsequent efforts have further emphasized robustness and expanded coverage, including SWE-bench Verified~\citep{swebenchverified} and SWE-bench++~\citep{swebenchpp}. Concurrently, other research initiatives have elevated ``repository operationalization'' to a first-class capability: Multi-Docker-Eval\citep{multidockereval} benchmarks agents on the automatic construction of runnable Docker environments across heterogeneous projects.

\textbf{Limitations.}
While these efforts substantially improve evaluation reproducibility, they predominantly frame containerization as an \textbf{evaluation harness} or a specific \textbf{benchmark task}. Consequently, environment setup remains tightly coupled to particular benchmark specifications, and the environment layer is rarely abstracted as a reusable, versioned asset capable of supporting diverse downstream task families at scale. Furthermore, failure handling in large-scale automation is frequently under-specified; distinguishing between environment failures, test harness incompatibilities, and invalid task instances remains a persistent challenge.

\textbf{Our improvement.}
SWE-Hub generalizes containerized evaluation into a shared \textbf{execution substrate} for data production. By converting repositories into versioned, pinned container images with standardized verifier entrypoints, we decouple the environment logic from task logic. Downstream generators consume only this uniform interface, enabling systematic gating, replay, and the reliable support of multiple task families over the same substrate.

\subsection{Scalable Synthesis and Task Generation}

Complementing static benchmarks, several lines of work aim to scale training and evaluation data by synthesizing verifiable tasks from repositories. SWE-Smith~\citep{swesmith} generates large volumes of bug-fix instances by manipulating real codebases and validating candidates via execution. Similarly, R2E-Gym~\citep{r2egym} and SWE-Gym~\citep{swegym} provide executable training environments and procedural task generation with verifiers to support agent learning under rigorous correctness checks. Other efforts explore alternative scaling axes. For instance, SWE-Synth~\citep{swesynth} synthesizes verifiable data with agent-generated trajectories, while SWE-Mirror expands datasets by reusing existing environments and ``mirroring'' issues across repositories~\citep{swe-mirror}. Additionally, Self-Play SWE-RL~\citep{selfplayswerl} automates problem creation via iterative self-play. Finally, SWE-Lego~\citep{swelego} demonstrates that carefully constructed mixtures of real and synthetic data can substantially improve supervised fine-tuning.

\textbf{Limitations.}
Most existing synthesis pipelines prioritize a single task format—often short-horizon repairs—and embed implicit assumptions about environment setup, repository structure, or verification interfaces. This coupling limits portability across ecosystems and hinders the composition of diverse task families. Additionally, while high-throughput synthesis requires treating verification as a systems problem (encompassing isolation, retry semantics, stateless execution, and scalable scheduling), many prior pipelines validate tasks without explicitly surfacing the infrastructure contract as a reusable layer.

\textbf{Our improvement.}
SWE-Hub provides a unified factory architecture where (i) environment provisioning is decoupled from task logic via a standardized execution substrate, and (ii) task generation is organized into modular product lines sharing a common verification contract. This design supports high-throughput synthesis and filtering through scalable, stateless sandboxed verification, while simultaneously enabling task diversity that extends beyond local bug-fixing.

\subsection{Long-Horizon Competencies and Repository Creation}
Recognizing that issue-fixing alone is insufficient to measure comprehensive agent capabilities, recent benchmarks increasingly emphasize long-horizon competencies such as planning, dependency management, and multi-step repository evolution. For instance, SWE-EVO evaluates coding agents in long-horizon software evolution scenarios~\citep{sweevo}, and SWE-Bench Pro introduces more complex tasks extending beyond standard issue fixing~\citep{swebenchpro}. The NL-to-repository setting further pushes the boundary by requiring agents to construct or substantially extend repositories from natural language requirements under executable tests, with NL2Repo-Bench serving as a prominent example~\citep{nl2repobench}.

\textbf{Limitations.}
While these benchmarks successfully motivate long-horizon evaluation, they are typically introduced as standalone datasets or isolated tasks. They lack integration into a shared production pipeline capable of continuously generating and validating instances under a uniform execution interface.

\textbf{Our improvement.}
SWE-Hub integrates long-horizon repository construction as a first-class product line built upon the same execution substrate used for repair and regression tasks. This integration ensures consistent verification, comparable task records, and scalable instance generation across different task horizons.

\subsection{Summary}
In summary, existing work has advanced (i) containerized reproducible evaluation and (ii) scalable synthesis of executable SWE tasks, yet these capabilities are often delivered as benchmark-specific harnesses or single-format generation pipelines. SWE-Hub contributes a production-oriented unification: a reusable execution substrate with standardized verifier interfaces, coupled with multiple task product lines that span the spectrum from short-horizon repairs and system-level regressions to long-horizon repository construction.

\begin{figure}[t]
    \centering
    \includegraphics[width=\linewidth]{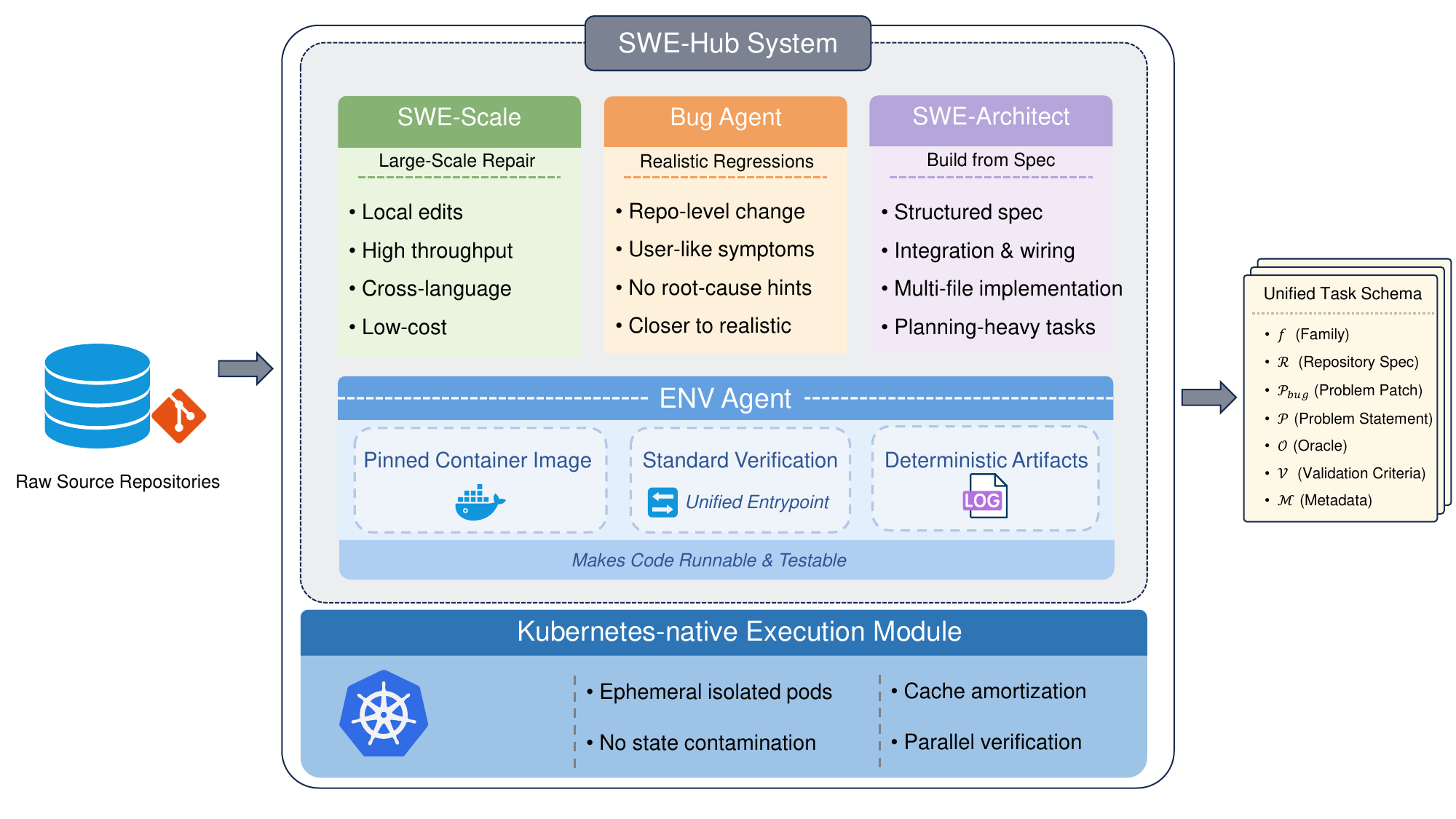} 
    \caption{
\textbf{SWE-Hub architecture.}
Starting from raw source repositories, the \textbf{Env Agent} provisions a deterministic execution substrate—pinned container images, a unified verification entrypoint, and deterministic artifacts—to make code runnable and testable.
On top of this substrate, three task generators produce complementary SWE workloads: \textbf{SWE-Scale} for high-throughput local repair edits, \textbf{Bug Agent} for realism-oriented repo-level regressions with user-like symptoms and no root-cause hints, and \textbf{SWE-Architect} for building multi-file implementations from structured specifications.
    }
    \label{fig:swe_hub}
\end{figure}

\section{SWE-Hub System Overview}
\label{sec:system_overview}

SWE-Hub is an end-to-end \emph{data production system} designed to generate diverse, \emph{executable} software engineering (SWE) tasks at scale. Given raw source repositories, SWE-Hub produces task instances that are \emph{verifiable by construction}: each instance is packaged with a pinned execution environment and a standardized verification entrypoint. Crucially, SWE-Hub represents not a static benchmark release, but a \emph{factory abstraction} that decouples (i) a shared \emph{execution substrate}—responsible for making repositories runnable and testable—from (ii) multiple \emph{task product lines} that synthesize distinct families of tasks on top of this substrate. Built upon a foundational Kubernetes execution environment, this architecture ensures the scalability and isolation necessary for high-throughput data production. This section details the design goals, system architecture, unified task schema, and the Kubernetes-native execution module.

\subsection{Design Goals}
\label{sec:design_goals}

We design SWE-Hub around five verifiable goals, each directly manifested in the system's artifacts and operational guarantees.

\begin{itemize}
    \item \textbf{G1: Executable by construction.} Every task is bound to (a) an immutable container image reference (pinned by digest) and (b) a standardized verification entrypoint (test command + result collector). This eliminates implicit environment assumptions and ensures that correctness is determined strictly through execution signals.
    \item \textbf{G2: Cross-language extensibility with bounded engineering effort.} SWE-Hub is language-agnostic at the system level. Language- and ecosystem-specific details (parsing rules, build conventions, dependency tooling, test runners) are externalized into declarative configurations. A shared analysis backbone (e.g., tree-sitter-based parsing and queries) provides a uniform interface for code inspection and transformation. This design does not claim a ``universal solution'' for all languages but significantly reduces the marginal cost of adding new ecosystems to implement the required configuration and adapters.
    \item \textbf{G3: Verification as a first-class scalability target.} Since execution dominates the computational cost, verification is treated as a distributed systems problem. SWE-Hub employs stateless, containerized workers scheduled on a Kubernetes cluster~\citep{kubernetes} to support high parallelism, strict isolation, and fault recovery. The system aims for throughput that scales linearly with available compute while remaining robust to long-tail test runtimes and I/O overheads.
    \item \textbf{G4: Realism with leakage-resistant task presentation.} Beyond simple local perturbations, SWE-Hub supports tasks that mirror real-world incidents: system-level regressions accompanied by user-like issue reports. Issue prompts are generated under explicit constraints that preserve observable symptom and reproduction evidence (logs, steps, expected/observed behavior) while withholding root-cause hints (e.g., causal identifiers and explanations), thereby mitigating answer leakage.
    \item \textbf{G5: Multi-horizon task coverage under a unified verifier interface.} The system supports both short-horizon repair tasks and long-horizon repository construction tasks. All task families are evaluated through the same execution substrate and verification entrypoint; task-specific oracles dictate only how success is interpreted (e.g., fail-to-pass sets vs. hidden tests), enabling comparable evaluation across different complexity horizons.

\end{itemize}

\subsection{Architecture}
\label{sec:architecture}

\paragraph{Overview.}
SWE-Hub is composed of an \emph{execution substrate layer} and multiple \emph{task product lines} (Figure~\ref{fig:swe_hub}). The substrate layer guarantees that a repository is \emph{runnable and testable} in a deterministic way, while the product lines generate task instances by transforming repositories and validating outcomes through execution.

\paragraph{Execution substrate: Env Agent.}
Given a repository snapshot, the \textbf{Env Agent} automatically constructs:
(i) a versioned container image capturing toolchains and dependencies, and
(ii) standardized verification entrypoints (e.g., a test runner command coupled with a result parser).
The target state is defined as environment readiness: the repository's verification harness can be invoked deterministically inside the container and produces machine-parseable outcomes (including failures), with logs and exit status captured.
Notably, readiness does not require tests to pass; rather, it establishes a stable boundary that ensures downstream synthesis and evaluation are reproducible.

\paragraph{Product line A: SWE-Scale for scalable repair-task synthesis.}
Building on the SWE-Smith methodology, \textbf{SWE-Scale} engine produces high volumes of short-horizon repair instances by injecting candidate fault patches and validating them through execution.
It leverages cross-language code analysis (via the shared parsing backbone) to locate transformation sites and apply either procedural mutations or LLM-assisted rewrites.
Candidates are accepted only if execution validates the intended signal (e.g., pass-to-fail transitions under the repository test suite), yielding verifiable repair tasks with precise failure signatures.

\paragraph{Product line B: Bug Agent for realism-oriented regressions and issue prompts.}
\textbf{Bug Agent} targets failures that are non-local in either origin or manifestation.
It synthesizes system-level regressions involving cross-file or cross-module interactions (e.g., contract violations across components, configuration-driven behavior changes, or regressions whose symptoms surface far from the modified code).
For each validated regression, Bug Agent generates a user-like issue report that includes symptoms and reproduction evidence while explicitly avoiding root-cause disclosure.
While Bug Agent shares the substrate and verifier interface, it operates independently to generate high-fidelity debugging scenarios.

\paragraph{Product line C: SWE-Architect for long-horizon repository construction.}
Building on the NL2Repo methodology, \textbf{SWE-Architect} expands the task space from repair to creation.
Starting from a \textbf{pristine repository version}, it derives structured natural-language requirements and generates tasks that require agents to implement substantial repository functionality (modules, APIs, integration glue) from specifications.
Evaluation remains execution-grounded under the same substrate and entrypoint, typically utilizing hidden tests to assess repository-scale correctness, architectural planning quality, dependency management, and global consistency.

\paragraph{Shared services.}
All product lines rely on common infrastructure: (i) the execution substrate (image + entrypoints), (ii) a unified task schema, (iii) cluster-scale orchestration and isolation, and (iv) standardized logging and artifact retention.
This modular design allows new task families to be added as additional product lines without re-engineering environment provisioning or verification logic.

\subsection{Unified Task Schema}
\label{sec:task_schema}

SWE-Hub records every task as a self-contained, language-agnostic task record optimized for consumption by downstream training and evaluation pipelines.
Formally, each task instance $T$ is defined as:
\[
T = (f, \mathcal{R}, \mathcal{P}_{\text{bug}}, \mathcal{P}, \mathcal{O}, \mathcal{V}, \mathcal{M}),
\]
where:

\begin{itemize}
    \item \textbf{Family} $f \in \{\text{SWE-Scale}, \text{Bug Agent}, \text{SWE-Architect}\}$: the task family produced by a specific product line.
    \item \textbf{Repository Spec} $\mathcal{R}$: repository identifier and pinned base commit representing the clean, original state.
    \item \textbf{Problem Patch} $\mathcal{P}_{\text{bug}}$: the patch applied to $\mathcal{R}$ to generate the initial (buggy or incomplete) repository state presented to the agent.
    \begin{itemize}
        \item For repair tasks ($f \in \{\text{SWE-Scale}, \text{Bug Agent}\}$), this is the injected bug patch.
        \item For creation tasks ($f = \text{SWE-Architect}$), this is the hollowing patch that removes implementations to define the starting point.
    \end{itemize}
    \item \textbf{Problem Statement} $\mathcal{P}$: a structured natural-language prompt (e.g., issue-style description) describing the task.
    \item \textbf{Oracle} $\mathcal{O}$: the \textbf{Ground Truth Solution} patch.
    \begin{itemize}
        \item For repair tasks, this is the \textbf{reference fix patch} that resolves the issue.
        \item For creation tasks, this is the \textbf{gold implementation patch} containing the correct modules and functions.
    \end{itemize}
    \item \textbf{Validation Criteria} $\mathcal{V}$: the specific test execution signals or partitions used to judge success, independent of the oracle.
    \begin{itemize}
        \item For repair tasks, $\mathcal{V}$ consists of the \textbf{test partition} derived from the baseline run, specifically identifying the \textbf{pass-to-fail} ($T_{\text{P2F}}$) and \textbf{pass-to-pass} ($T_{\text{P2P}}$) test sets that define the required behavior change.
        \item For creation tasks, $\mathcal{V}$ consists of the \textbf{target test suite} (hidden tests) that must be satisfied by the completed repository.
    \end{itemize}    
    \item \textbf{Metadata} $\mathcal{M}$: execution-grounded diagnostics and performance statistics, including exit codes, runtime/resource usage, and full logs (stdout/stderr).
\end{itemize}

This schema is intentionally \textbf{execution-centered}: it separates the initial state construction ($\mathcal{R} + \mathcal{P}_{\text{bug}}$) from the solution verification. The oracle $\mathcal{O}$ strictly represents the correct code, while $\mathcal{V}$ defines the behavioral contract the solution must satisfy. In practice, all families reduce to invoking the same verification entrypoint inside the pinned image. Here, the field $\mathcal{V}$ governs the automated evaluation by determining success based solely on execution signals (e.g., unit test pass/fail status), whereas $\mathcal{O}$ serves as the ground-truth reference code for offline analysis and training, distinct from the execution-based verification process.

\subsection{Kubernetes-native Execution Module}
\label{sec:execution_model}

SWE-Hub is implemented as a container-native distributed system to maximize throughput, ensure isolation, and guarantee operational robustness.

\paragraph{Job orchestration.}
A central controller schedules environment provisioning, task generation, and verification as Kubernetes Jobs. Each unit of work executes in a short-lived pod with explicit CPU/memory limits, preventing state contamination across tasks and enforcing strict resource boundaries.

\paragraph{Stateless verification workers.}
Verification runs are designed to be stateless: inputs are task records and pinned image references, while outputs are structured validation artifacts.
This statelessness supports deterministic reruns, robust retry-on-transient-failure policies, and simplified debugging.

\paragraph{Caching and warm start.}
To mitigate overhead, cluster nodes cache commonly used images and dependencies.
For repositories with expensive setup costs (e.g., complex builds), SWE-Hub can maintain a warm container pool to amortize initialization time while strictly preserving isolation guarantees.

\paragraph{Observability and artifact retention.}
For every execution, SWE-Hub logs comprehensive stdout/stderr, exit codes, parsed test outcomes, and resource usage.
Failures are retained with full diagnostic context to enable offline analysis and to prevent silent dataset corruption.

\begin{figure}[htbp]
    \centering
    \includegraphics[width=\linewidth]{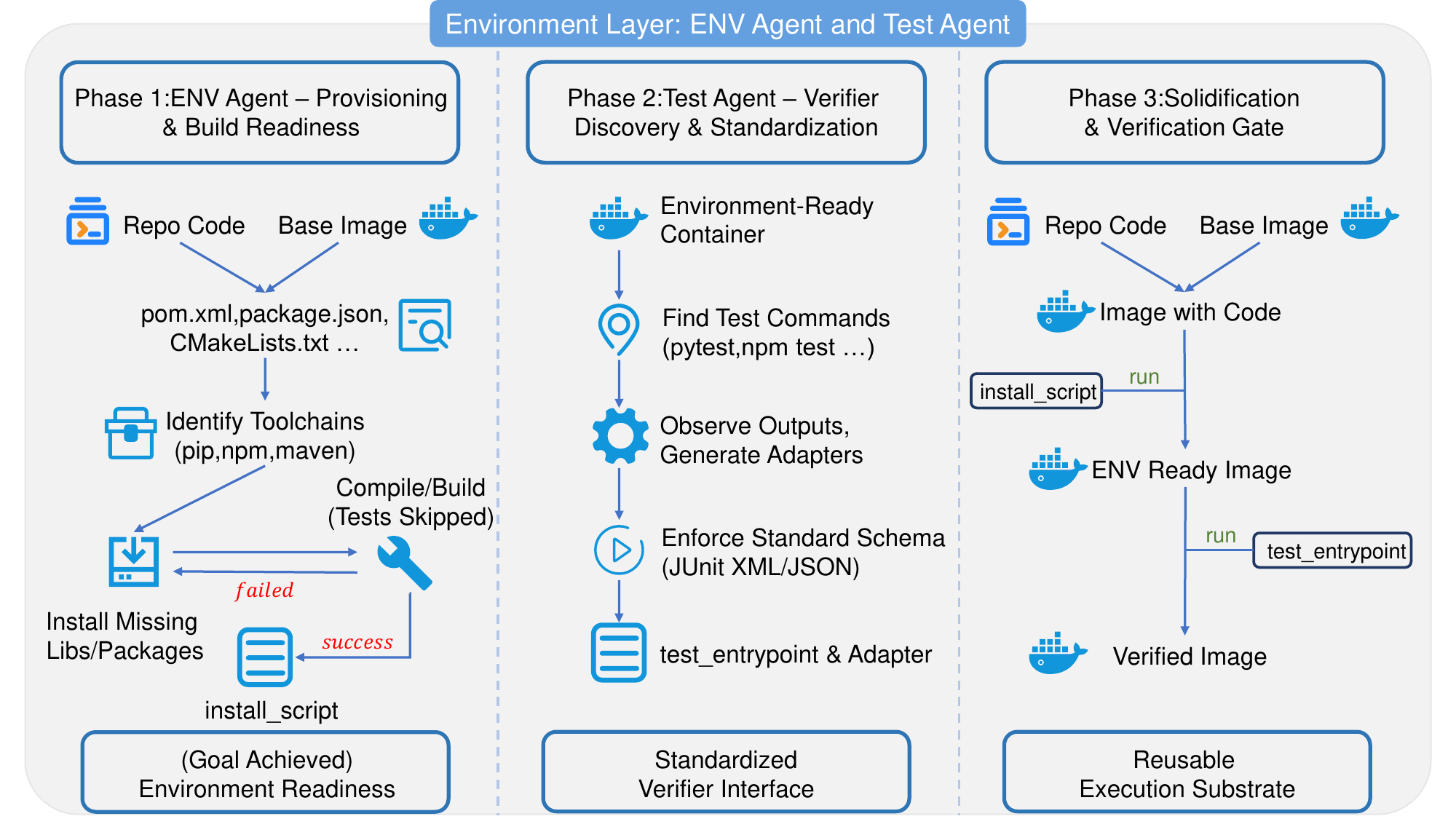} 
    \caption{
        \textbf{SWE-Hub Environment Layer pipeline.} \textbf{Phase 1 (Env Agent)} performs environment provisioning and build readiness, identifying toolchains and installing dependencies to achieve \textit{environment readiness} without requiring tests to pass. \textbf{Phase 2 (Test Agent)} establishes a unified verification interface via entrypoint discovery and result standardization, employing dynamic adapters to handle heterogeneous ecosystems where native reporting is non-standard. \textbf{Phase 3 (Solidification and Verification Gate)} encapsulates the environment state and verification logic into a pinned container image, applying a strict execution check to ensure the substrate is deterministic and reusable for downstream task generation.
    }
    \label{fig:env_agent}
\end{figure}

\section{Environment Layer: The Execution Substrate}
\label{sec:env_layer}

Repository-level software engineering is fundamentally \emph{execution-grounded}: the success of an agent is ultimately determined by build correctness, dependency resolution, runtime behavior, and executable test suites. In practice, the primary source of brittleness when scaling training data synthesis and evaluation is not the code transformation logic itself, but the \emph{environment problem}: heterogeneous toolchains, drifting dependencies, project-specific test entrypoints, and non-standard result formats. SWE-Hub addresses this bottleneck by elevating environment provisioning to a first-class component of the factory. Our Environment Layer deploys two specialized agents—\textbf{Env Agent} and \textbf{Test Agent}—to transform an arbitrary repository snapshot into a reusable \emph{execution substrate} that downstream task product lines can invoke deterministically(Figure~\ref{fig:env_agent}).

\paragraph{Outputs.}
For each repository snapshot, the Environment Layer produces three core artifacts:
(i) an idempotent \texttt{install\_script} that provisions toolchains and dependencies;
(ii) a standardized \texttt{test\_entrypoint} that executes the repository's verification harness and emits machine-parseable results; and
(iii) a versioned container image that encapsulates the above into a durable, reusable asset.
Downstream components (SWE-Scale, Bug Agent, SWE-Architect) consume this substrate via a uniform interface: \emph{invoke the \texttt{test\_entrypoint} inside the pinned image and parse the standardized artifacts}.

\paragraph{Section contributions.}
This section makes the following contributions: (i) \textbf{Formalization of environment readiness.} We define a strict termination condition separating infrastructure setup (invocability) from test passing, resolving the ambiguity between environment failures and test failures. (ii) \textbf{Dual-agent framework.} We introduce a decoupled Env Agent and Test Agent design to automate complex dependency resolution and test harness discovery across diverse ecosystems. (iii) \textbf{Dynamic adapters.} We provide runtime adapters to normalize heterogeneous test outputs into a unified schema, ensuring consistent downstream verification.

\subsection{Challenges and Design Principles}
\label{sec:env_principles}

A single ``run tests'' command is insufficient as a robust objective across diverse ecosystems. In many repositories, failures occurring during environment setup (e.g., missing system libraries, incompatible runtime versions, broken dependency resolution, or absent build tools) are indistinguishable from legitimate test failures (e.g., assertion failures signaling real bugs) if conflated into a single execution stage. This ambiguity is fatal for automation: it prevents the system from determining whether it should continue remedying the environment or proceed to downstream task generation.

SWE-Hub therefore adopts two core principles:
\begin{itemize}
    \item \textbf{P1: Decouple \emph{environment readiness} from \emph{test correctness}.} The Environment Layer is responsible solely for making the verification harness invocable and its outputs collectible; it is explicitly \emph{not} responsible for ensuring that tests pass.
    \item \textbf{P2: Standardize the verifier interface and artifacts.} To support large-scale distributed execution, verification must rely on a stable entrypoint and a canonical output schema, regardless of the underlying language, framework, or project-specific conventions.
\end{itemize}

These principles motivate a two-agent design: the \textbf{Env Agent} establishes readiness by provisioning a buildable, runnable environment, while the \textbf{Test Agent} identifies test entrypoints and produces machine-parseable results—often implementing adapters when native structured reports are unavailable.

\subsection{Environment Readiness as a Termination Criterion}
\label{sec:readiness}
A central requirement for autonomous environment provisioning is a clear, verifiable termination condition. We define \textbf{environment readiness} as a repository-specific boundary state at which:
\begin{quote}
\emph{the repository's verification harness can be invoked deterministically inside the container, and the run produces machine-parseable outcome artifacts (including failures), with logs and exit status captured.}
\end{quote}

Readiness explicitly does \emph{not} imply that ``tests pass.'' Rather, it signifies that the system can repeatedly execute the verifier and collect structured outcomes. Concretely, readiness is instantiated per ecosystem by verifying the presence of a valid test invocation capability coupled with structured artifact generation:
\begin{itemize}
    \item \textbf{Python}: a \texttt{pytest}-based harness is invocable and emits a structured report (e.g., JUnit XML or JSON), regardless of individual test pass/fail status.
    \item \textbf{Java}: \texttt{mvn test} or \texttt{gradlew test} executes the test harness to completion and produces JUnit-compatible XML reports.
    \item \textbf{JS/TS}: a project-defined test script executes without environment-level errors (e.g., missing modules or toolchains), and its results are captured in a standardized schema (either native or adapted).
    \item \textbf{C\# / .NET}: \texttt{dotnet test} executes against the solution or project targets and emits structured results (e.g., TRX format via built-in loggers) regardless of test pass/fail status.
    \item \textbf{C/C++}: the project builds and runs test targets (via CMake, Make, or Autotools), and the verifier (e.g., \texttt{ctest} or test executables) is invocable with logs and exit codes captured. Execution outputs are normalized into a unified schema, utilizing native reports or generated adapters where necessary.
    \item \textbf{Go}: \texttt{go test} executes across relevant packages (e.g., \texttt{./...}), building dependencies and test binaries, and emits structured JSON output (via the \texttt{-json} flag) capturing individual test status.
    \item \textbf{PHP}: dependencies are resolved via Composer, and the test harness (e.g., PHPUnit via \texttt{vendor/bin/phpunit}) runs to completion, generating structured logs (e.g., JUnit XML) regardless of pass/fail status.
    \item \textbf{Ruby}: the environment is prepared using Bundler, and the test suite (e.g., RSpec or Minitest via \texttt{bundle exec}) executes. Results are captured in a standardized schema, using native JSON formatters or adapters to normalize output streams.
    \item \textbf{Rust}: the project builds and tests via \texttt{cargo test}, which implicitly compiles the workspace. The test harness runs to completion, and results are captured—typically via an adapter that parses the standard output or through specific logging configurations—to produce machine-parseable artifacts.
\end{itemize}

This definition establishes a uniform, checkable success state while strictly respecting ecosystem-specific conventions.

The implementation of the Environment Layer follows the three-phase pipeline illustrated in Figure~\ref{fig:env_agent}: \textbf{Phase 1} (ENV Agent) establishes the build environment; \textbf{Phase 2} (Test Agent) standardizes the verification interface; and \textbf{Phase 3} solidifies these into a reusable image asset.

\subsection{Phase I: The Env Agent: Provisioning and Build Readiness}
\label{sec:env_agent}

\paragraph{Goal.}
The Env Agent generates an idempotent \texttt{install\_script} that transitions the container from a clean base image to an environment-ready state, without requiring tests to pass. Its output is a deterministic sequence of commands sufficient to reproduce the exact toolchain setup and dependency installation.

\paragraph{Workflow.}
Given a repository snapshot, the Env Agent executes the following steps:
\begin{enumerate}
    \item \textbf{Repository Scouting.} Identifies language and toolchain signals, locates build descriptors (e.g., \texttt{pyproject.toml}, \texttt{requirements.txt}, \texttt{package.json}, \texttt{pom.xml}, \texttt{build.gradle}), and infers likely runtime constraints (e.g., specific Node, Python, or Java versions).
    \item \textbf{Toolchain Provisioning.} Installs and pins the required runtimes and system dependencies (compilers, build tools, OS libraries) using non-invasive, environment-level operations.
    \item \textbf{Dependency Installation.} Executes ecosystem-appropriate installation commands (e.g., \texttt{pip}, \texttt{npm}, \texttt{maven}, \texttt{gradle}) while capturing detailed logs and exit codes.
    \item \textbf{Build Readiness Verification.} Executes build steps designed to avoid conflating test failures with environment failures (e.g., performing build/compile/package steps with tests skipped where supported). Failures trigger targeted remediation strategies (e.g., installing missing libraries, adjusting runtime versions, or installing build tools) until the readiness conditions are satisfied.
\end{enumerate}

\paragraph{Scope and Boundary.}
The Env Agent does not modify project semantics to ``fix tests'', nor does it attempt to interpret assertion-level failures. Its responsibility terminates once the repository can invoke the verifier deterministically.

\subsection{Phase II: The Test Agent: EntryPoint Discovery and Standardization}
\label{sec:test_agent}

\paragraph{Goal.}
Operating in an environment-ready container, the \textbf{Test Agent} produces a standardized \texttt{test\_entrypoint} (and any necessary adapters) such that a verification run yields machine-parseable artifacts under a uniform schema. This is critical because repository test commands and output formats are often project-specific and designed for human consumption.

\paragraph{Workflow.}
The Test Agent performs the following:
\begin{enumerate}
    \item \textbf{EntryPoint Discovery.} Identifies candidate test commands from project configuration files (e.g., \texttt{package.json} scripts, build tool defaults, CI configurations).
    \item \textbf{Trial Execution.} Executes candidate commands to observe failure modes and output structure, carefully distinguishing environment-level failures (missing tools, missing modules, misconfigured paths) from legitimate test failures (e.g., assertion errors) by parsing error signatures and exit codes.
    \item \textbf{Structured Reporting.} Prioritizes native structured reports where available (e.g., JUnit XML from Java toolchains, pytest JUnit/JSON plugins). When outputs are unstructured, it synthesizes an \textbf{adapter} that wraps the original test command and converts raw stdout/stderr streams into a unified \texttt{standard\_result.json} format.
    \item \textbf{Acceptance Validation.} The verifier is accepted if it can be invoked deterministically and produces valid structured artifacts, even if the tests themselves fail.
\end{enumerate}

\paragraph{Dynamic adapters for heterogeneous outputs.}
Adapters are particularly vital for ecosystems where testing conventions are highly heterogeneous or lack standardized reporting, such as JavaScript/TypeScript (characterized by a fragmented landscape of competing frameworks like Jest, Mocha, and AVA) and C/C++ (which often rely on custom build scripts and unstructured console output). In such cases, the Test Agent generates a lightweight runtime-specific wrapper (e.g., a Node.js runner for JS projects or a shell script adapter for C/C++ makefiles) that executes the project’s test script, captures output streams, and emits a normalized result object (containing test counts, failing test identifiers, exit codes, and logs). This ensures downstream verification remains uniform and machine-checkable across diverse repository ecosystems.

\subsection{Phase III: Substrate Solidification: From Scripts to Images}
\label{sec:solidify}
While the Env Agent and Test Agent initially operate in an exploratory mode to generate per-repository scripts and adapters, SWE-Hub \emph{solidifies} these ephemeral artifacts into durable, reusable assets:
\begin{itemize}
    \item A \textbf{versioned container image} produced by applying the \texttt{install\_script} to a clean base image and freezing the resulting filesystem state.
    \item A \textbf{metadata record} that binds the image reference to the specific verifier interface (test entrypoint + result schema) utilized throughout the factory.
\end{itemize}

\paragraph{Verification gate.}
Before an image enters the asset pool, SWE-Hub enforces a strict validation gate: it instantiates a fresh container from the candidate image and executes the standardized \texttt{test\_entrypoint}. The environment is accepted only if the run completes deterministically and emits valid machine-parseable artifacts conforming to the expected schema (with logs archived). This gate prevents silent corruption of downstream datasets and ensures that all subsequent synthesis and evaluation steps inherit strong reproducibility guarantees.

\begin{figure}[t]
    \centering
    \includegraphics[width=\linewidth]{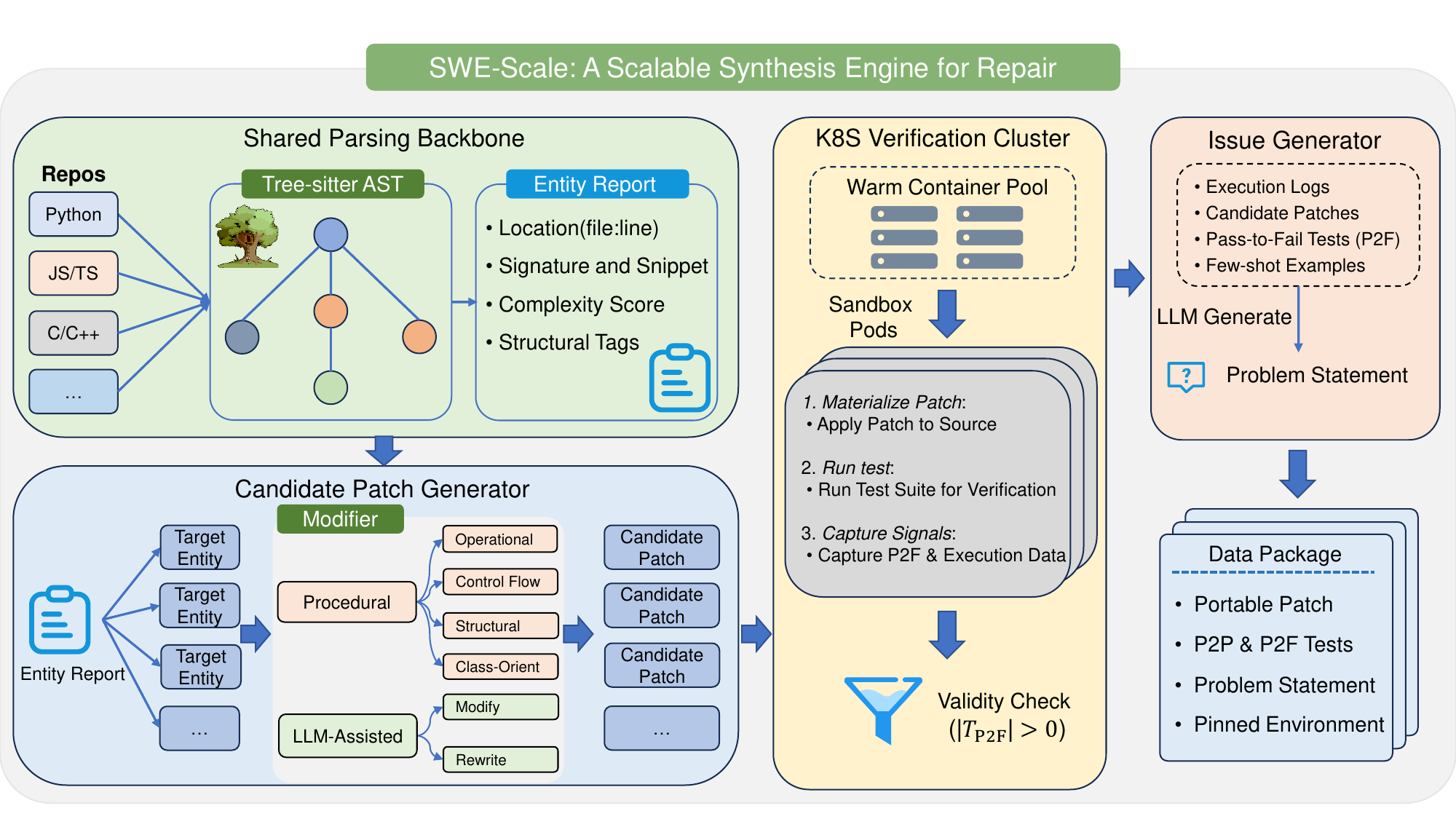} 
    \caption{
        \textbf{SWE-Scale architecture.} The system transforms an \textbf{Environment-Ready Substrate} into executable bug-fix instances. The pipeline follows a structured flow: the \textbf{Shared Parsing Backbone} performs polyglot entity analysis using Tree-sitter ASTs to identify targets; the \textbf{Candidate Generator} synthesizes fault specifications via procedural or LLM-assisted strategies; and the \textbf{Kubernetes-Scale Verification Cluster} materializes patches in isolated sandbox pods, executing tests and performing behavioral validity checks to produce verified task records.
    }
    \label{fig:swesmith_arch}
\end{figure}

\section{SWE-Scale: A Scalable Synthesis Engine for Repair}
\label{sec:swesmith}

SWE-Hub’s primary source of data volume is \textbf{SWE-Scale}, a high-throughput synthesis-and-verification engine that transforms a single \emph{environment-ready} repository substrate into vast numbers of \emph{executable} bug-fix instances. The critical bottleneck in prior synthesis workflows is shared mutable state: when generation and validation operate on a single working directory (or a shared Git checkout), each candidate mutation forces a serial loop of apply $\rightarrow$ test $\rightarrow$ reset. SWE-Scale eliminates this bottleneck by enforcing three system-level invariants: (\emph{i}) \textbf{stateless sandboxes} for all mutations and tests, (\emph{ii}) \textbf{cluster-scale verification} as the default execution mode, and (\emph{iii}) \textbf{polyglot synthesis} via a shared parsing backbone and configuration-driven language support.

\paragraph{Section contributions.}
This section makes three contributions aligned with these invariants:
(i) a container-native redesign that enables massively parallel verification,
(ii) a polyglot analysis-and-mutation substrate grounded in a unified parser and configuration-as-code, enabling cross-language bug injection without core engine rewrites, and
(iii) a restructured bug-injection engine that emits robust, byte-offset edits and hosts both procedural and model-assisted transformations.
Figure~\ref{fig:swesmith_arch} illustrates the overall architecture of SWE-Scale, integrating these components into a unified pipeline.

\subsection{Stateless Execution Module and Container Isolation}
\label{sec:swesmith_stateless}

SWE-Scale is architected around a strict ``sandbox-per-candidate'' execution model. Leveraging the pinned container image and verifier entrypoint produced by the Environment Layer (Section~\ref{sec:env_layer}), each atomic operation (e.g., materializing a candidate edit, extracting a patch, or executing a verification job) is run inside a fresh, short-lived container instance.
No candidate mutates the global checkout; instead, each receives an isolated filesystem snapshot and is discarded immediately after artifact generation.

This design yields two critical benefits.
\textbf{First}, it eliminates cross-candidate contamination (e.g., lingering build products, modified lockfiles, cached dependencies) that can silently corrupt validation results.
\textbf{Second}, it ensures failures are local and recoverable: a crashed container is equivalent to a failed candidate and can be retried or dropped without affecting concurrent jobs.
In practice, this statelessness is the prerequisite for scalable verification (Section~\ref{sec:swesmith_k8s}) and for producing reproducible task artifacts that downstream pipelines can deterministically replay.

\paragraph{Warm-start optimization via a container pool.}
While stateless sandboxes guarantee correctness and isolation, naive container instantiation can impose significant startup overhead.
SWE-Scale therefore maintains a \emph{warm container pool} that leases pre-initialized sandboxes (with the repository snapshot and common tooling pre-loaded), applies candidate edits, and returns sandboxes after cleanup.
This mechanism amortizes cold-start costs without compromising the stateless execution contract.

\subsection{Distributed Verification and Behavioral Validity Criteria}
\label{sec:swesmith_k8s}

Verification is the dominant computational cost in synthesis pipelines. SWE-Scale therefore treats test execution as a distributed systems workload: candidate patches are dispatched as Kubernetes jobs, each executing within an isolated pod subject to explicit resource limits and standardized logging. This architecture enables high concurrency while preserving reproducibility through pinned images and uniform verifier entrypoints.

\paragraph{Behavioral validity criteria.}
Let $T_{\text{suite}}$ denote the repository’s test suite, and let $R(P, \mathcal{E})$ be the vector of test outcomes obtained by applying patch $P$ in environment $\mathcal{E}$ and invoking the standardized verifier entrypoint. 
For a candidate injected patch $P_{\text{bug}}$ to be accepted as a valid regression instance, it must flip at least one previously passing test to failing (a non-empty pass-to-fail set):
\[
T_{\text{P2F}} = \{ t \in T_{\text{suite}} \mid R(\emptyset,\mathcal{E})[t]=\texttt{pass} \wedge R(P_{\text{bug}},\mathcal{E})[t]=\texttt{fail} \}, \quad |T_{\text{P2F}}| > 0.
\]
We also record the pass-to-pass set $T_{\text{P2P}}$ for regression analysis and stability checks:
\[
T_{\text{P2P}} = \{ t \in T_{\text{suite}} \mid R(\emptyset,\mathcal{E})[t]=\texttt{pass} \wedge R(P_{\text{bug}},\mathcal{E})[t]=\texttt{pass} \}.
\]

These execution-grounded signals, combined with logs, exit codes, runtime metrics, and resource usage, are stored in the task Validation field $\mathcal{V}$ and Metadata field $\mathcal{M}$ (Section~\ref{sec:task_schema}), enabling comprehensive auditing, deterministic reruns, and the extraction of rich downstream training signals beyond a single binary label.

\subsection{Polyglot Analysis and Targeting Backbone}
\label{sec:swesmith_polyglot}

Earlier synthesis systems are often tightly coupled to specific language ecosystems (e.g., Python AST tooling), making extensibility expensive. SWE-Scale adopts a polyglot design centered on \textbf{tree-sitter} as a shared parsing and querying backbone, which provides a uniform interface for analyzing and transforming syntactic structures across languages.

\paragraph{Universal entity extraction.}
The first phase of the synthesis pipeline transforms raw source code into a structured, language-agnostic representation. By utilizing the shared tree-sitter backbone~\citep{tree-sitter}, the system identifies and extracts code entities—such as functions, methods, and classes—irrespective of the programming language. This analysis generates an \emph{entity analysis report} that captures essential structural features (e.g., signatures, nesting depth) and complexity signals (e.g., presence of loops or binary operations). This report serves as the foundational inventory for the engine, enabling precise targeting and filtering of candidates based on their structural properties.

\paragraph{Enabling procedural modification.}
Beyond analysis, the shared tree-sitter backbone also serves as the enabler for cross-language bug injection. It allows the engine to precisely \emph{locate} modification targets in any supported ecosystem without relying on language-specific AST libraries. By leveraging the same querying mechanism used for extraction, the engine can identify specific syntactic patterns (e.g., binary operators, conditional blocks) that serve as injection points. This capability ensures that the logic for target localization is decoupled from the specific syntax of the source language.

\paragraph{Configuration-as-code for language support.}
To achieve true polyglot extensibility without rewriting the core engine, SWE-Scale externalizes all language-specific knowledge into declarative \textbf{Language Templates} (YAML). These templates define the rules governing both analysis and modification:
\begin{itemize}
    \item \textbf{Metadata}: File extensions, indentation size, and test file patterns.
    \item \textbf{Analysis Rules}: Patterns used to identify entities and compute complexity.
    \item \textbf{Injection Rules}: Queries that map bug types to syntactic structures (e.g., locating binary expressions for operand swaps).
\end{itemize}
This design ensures that supporting a new language (e.g., Java, JavaScript, or Go) requires only authoring a new YAML configuration, allowing the system to seamlessly scale its synthesis capabilities across diverse codebases.

\subsection{Modification Strategies: Procedural and LLM-Based}
\label{sec:swesmith_modify}

Built upon the shared parsing interface, SWE-Scale implements bug injection as a set of pluggable transformation policies. The engine supports two distinct modes of fault generation: \textbf{Procedural Modification} (rule-based structural edits) and \textbf{LLM-Based Modification} (semantic-level rewrites).

\subsubsection{Procedural Modification: Rule-Based Structural Edits}
\label{sec:swesmith_procedural}

Building upon the targeting capabilities of the polyglot backbone, procedural modification implements a standardized, three-step workflow for injecting syntax-level faults. \textbf{First}, the engine \textbf{localizes} target nodes within the code using the tree-sitter queries defined in the Language Templates. \textbf{Second}, it validates \textbf{preconditions} against the entity analysis report to ensure the code is a suitable candidate (e.g., verifying that a function actually contains a loop before attempting a loop-modifying injection). \textbf{Finally}, it generates \textbf{atomic edits} based on the precise byte offsets of the located nodes. This approach guarantees that modifications are syntactically correct and structurally precise, avoiding the brittleness of naive string replacement.

We categorize these rule-based modifiers into four primary families, covering a wide spectrum of common programming errors:

\paragraph{Operational Modifications.}
These inject subtle arithmetic or logic errors by mutating operators or operands. Examples include swapping operators within the same functional group (e.g., changing $+$ to $-$), flipping logical operators (e.g., $==$ to $!=$), or swapping the order of operands in non-commutative operations (e.g., $a - b$ to $b - a$).

\paragraph{Control Flow Modifications.}
These disrupt the execution order or causal dependencies within functions. Techniques include inverting the bodies of \texttt{if-else} blocks, swapping \texttt{if-elif} conditions, or shuffling the order of statements within a function body to break temporal coupling.

\paragraph{Structural Modifications.}
These remove or unwrap code structures to simulate missing features or broken error handling. Examples include deleting loop bodies, removing conditional branches, stripping class inheritance to break polymorphism, or unwrapping \texttt{try-except} blocks to eliminate error handling.

Table~\ref{tab:procedural_modifiers} provides a comprehensive taxonomy of the implemented modifiers, detailing their target syntax, transformation logic, and the simulated bug types.

\subsubsection{LLM-Based Modification: Semantic Rewrites}
While procedural bugs are syntactically precise, they often lack the semantic nuance of real-world errors. SWE-Scale uses LLMs to generate two types of semantic bugs:

\paragraph{LLM Rewrite (Implementation Reconstruction).}
This mode simulates errors common in initial implementations.. The engine provides the LLM with the full file context but \textbf{obscures the implementation} of a target function (replacing it with a stub). The prompt instructs the model to implement the function from scratch without external libraries. This forces the model to rely on ``common sense'' patterns, often missing edge cases (e.g., null inputs, empty lists) or choosing inefficient algorithms, resulting in fragile code.

\paragraph{LLM Modify (Blind Implementation).}
This mode simulates ``maintenance-introduced'' errors. The engine provides the LLM with the \textbf{correct, working code} and instructs it to introduce a subtle logical bug that will break existing unit tests. The prompt provides a dynamic menu of bug types (e.g., ``introduce off-by-one error'', ``alter operator precedence'') and enforces strict constraints: no syntax errors, no comments revealing the bug, and high subtlety. This creates adversarial bugs that are harder to detect than simple syntactic flips.

\subsection{Issue Generation: Synthesizing User-Like Problem Statements}
\label{sec:swesmith_issue}

A complete task instance requires not only a patch but also a problem statement that guides the agent. After a valid bug is verified, SWE-Scale synthesizes an \textbf{Issue Description} that mimics real-world user reports.

\paragraph{Symptom-centric generation.}
The generation process is designed to translate technical failure artifacts into natural-language narratives that reflect a user's perspective, rather than a developer's debugging summary. This transformation is driven by three key inputs:
\begin{itemize}
\item \textbf{The Bug Patch}: The specific code changes that introduced the regression.
\item \textbf{Execution Artifacts}: The failure traces (stack traces, error messages, and stdout) which provide the observable symptoms of the bug.
\item \textbf{Failing Test Source}: The source code of the test cases that failed ($T_{\text{P2F}}$), which defines the expected behavior and context.
\end{itemize}
The LLM constructs a symptom-focused narrative by analyzing the test source code to understand expected behavior and contrasting it with the observed symptoms in execution logs. Prompted to act as a user, it describes the error and provides a reproduction script. To ensure the output captures the informal style of real community reports, the generation utilizes few-shot demonstrations from actual GitHub issues.

\subsection{End-to-End Workflow and Produced Artifacts}
\label{sec:swesmith_workflow}

Given an environment-ready repository substrate $(\mathcal{E})$, SWE-Scale executes a four-stage, fully parallelizable workflow:
\begin{enumerate}
    \item \textbf{Candidate enumeration and proposal.} The engine enumerates candidate entities using the shared parsing backbone and entity analysis report. It then generates candidate bug specifications using a hybrid of \textbf{procedural transformations} (Section~\ref{sec:swesmith_modify}) and \textbf{LLM-based modifications}.
    \item \textbf{Patch materialization in sandboxes.} Each candidate specification is dispatched to a disposable container (typically leased from the warm pool). Inside the sandbox, the candidate edit (whether byte-offset or LLM-generated code) is applied to the source, and a portable patch artifact (e.g., \texttt{git diff}) is extracted.
    \item \textbf{Distributed verification and filtering.} The candidate patch is verified in an isolated Kubernetes pod: apply patch $\rightarrow$ run standardized verifier $\rightarrow$ parse results. Only candidates satisfying the behavioral validity criterion (non-empty $T_{\text{P2F}}$) are retained.
    \item \textbf{Task packaging and Issue synthesis.} The final output of SWE-Scale is a verified \textbf{Data Package}, instantiating the task family $f = \text{SWE-Scale}$ within the unified task schema (Section~\ref{sec:task_schema}). For each retained candidate, the engine synthesizes a user-like problem statement $\mathcal{P}$ and aggregates it with the \textbf{repository spec $\mathcal{R}$}, the injected bug patch $\mathcal{P}_{\text{bug}}$, the oracle $\mathcal{O}$ (the reference fix patch), the validation criteria $\mathcal{V}$ (the $T_{\text{P2F}}$ and $T_{\text{P2P}}$ partitions), and execution metadata $\mathcal{M}$. This package is bound to the pinned execution environment $\mathcal{E}$, finalizing a self-contained, executable artifact ready for downstream consumption.
\end{enumerate}

\subsection{Discussion: Quality, Triviality, and System-Level Limitations}
\label{sec:swesmith_discussion}

Execution-based filtering ensures instances are active, but activeness is insufficient for high-quality data: trivial mutations (e.g., flipping a comparison) satisfy validity criteria yet offer limited learning signal. Moreover, entity-local transformations are inherently biased toward localized failures, failing to capture \textbf{system-level regressions} where symptoms are separated from their root causes by architectural layers.

These limitations drive two key implications. First, scaling synthesis throughput requires robust environment automation (Section~\ref{sec:env_layer}). Second, achieving system-level realism necessitates reasoning beyond local edits. To address this, SWE-Hub introduces \textbf{Bug Agent} (Section~\ref{sec:bug_agent}), an independent product line that targets complex, cross-module regressions and generates realistic, symptom-centric issue reports on the shared execution substrate.

\begin{table*}[htbp]
\centering
\caption{Taxonomy of Procedural Modifiers in SWE-Scale.}
\label{tab:procedural_modifiers}
\resizebox{0.9\textwidth}{!}{%
\begin{tabular}{@{}llp{6.5cm}l@{}}
\toprule
\textbf{Category} & \textbf{Modifier} & \textbf{Action} & \textbf{Bug Type} \\ \midrule
\textbf{Operational} 
 & Op. Change & Replace operator with group sibling (e.g., $+$ $\to$ $-$). & Operator Misuse \\
 & Op. Flip & Invert boolean/comparison logic (e.g., $>$ $\to$ $\le$). & Logic Inversion \\
 & Operand Swap & Swap L/R operands ($a,b \to b,a$). & Order Error \\
 & Chain Break & Replace full expression with sub-operand ($a+b \to a$). & Data Loss \\
 & Const. Mod & Apply $\pm 1$ to integer literals. & Off-by-one \\ \midrule
\textbf{Control Flow} 
 & Branch Swap & Swap \texttt{if} body with \texttt{else}/\texttt{elif} body. & Logic Reversal \\
 & Stmt Shuffle & Reorder statements within function scope. & Temporal Coupling \\ \midrule
\textbf{Structural} 
 & Block Drop & Remove entire Loop, Conditional, or Assignment. & Missing Logic \\
 & Wrap. Drop & Remove \texttt{try}/\texttt{with} block and its content. & Missing Feature \\
 & Wrap. Unwrap & Remove \texttt{try}/\texttt{with} keyword, \textbf{keep content}. & Unsafe Execution \\ \midrule
\textbf{Class-Oriented} 
 & Base Drop & Remove inherited base classes. & Broken Polymorphism \\
 & Mem. Shuffle & Reorder methods/fields within a class. & Declaration Order \\
 & Method Drop & Remove method definition. & Missing Interface \\ \bottomrule
\end{tabular}%
}
\end{table*}

\begin{figure*}[t] 
    \centering
    \includegraphics[width=\linewidth]{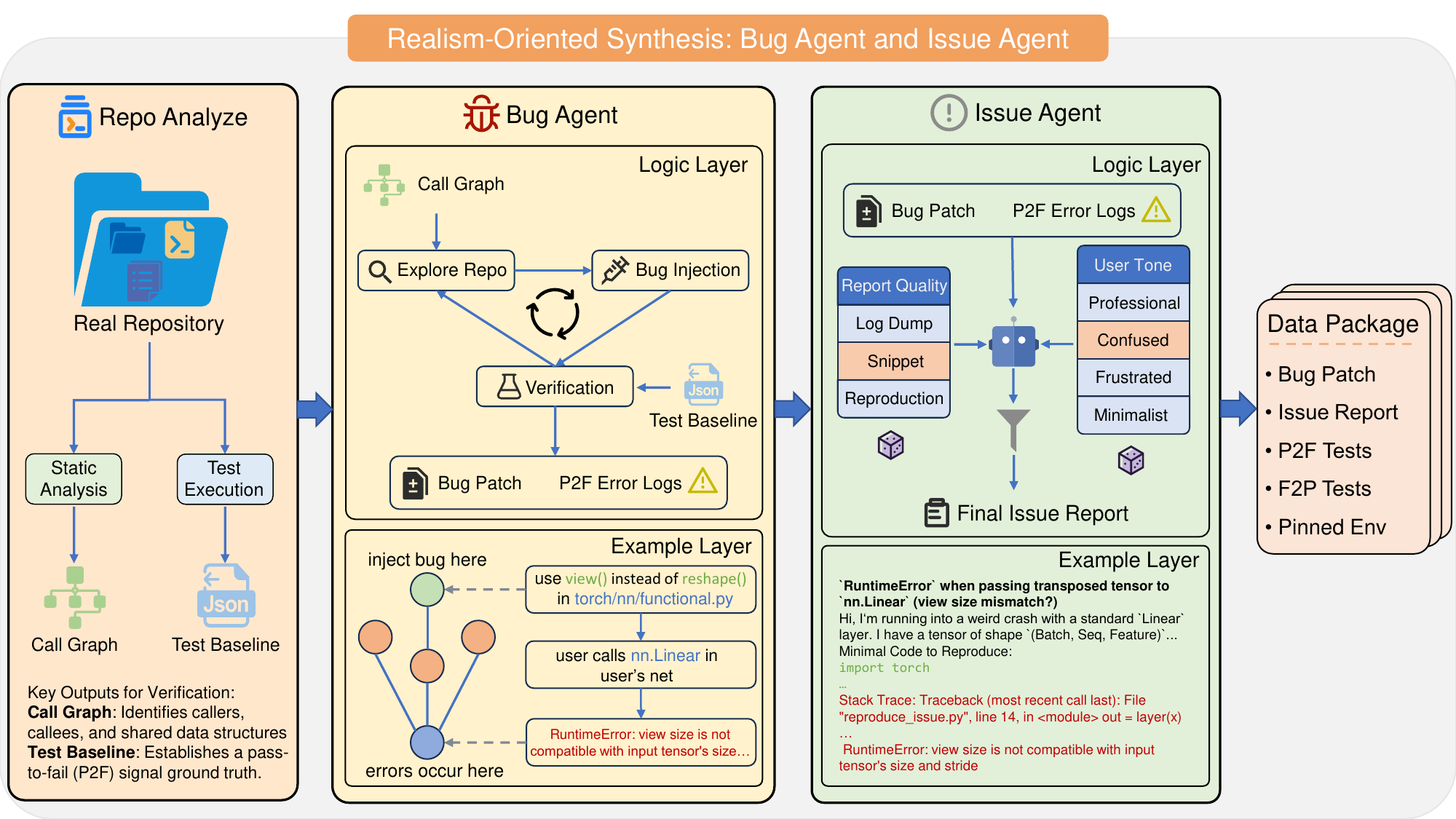} 
    \caption{
    \textbf{Workflow of the Realism Layer in SWE-Hub.} The pipeline initiates with \textbf{Repo Analysis} to extract call graphs and test baselines. The \textbf{Bug Agent} (center) then iteratively synthesizes system-level regressions that trigger non-local failures, validated by a Pass-to-Fail (P2F) signal. Finally, the \textbf{Issue Agent} (right) generates a user-like report from the verified regression, enforcing strict symptom-cause separation to prevent information leakage while adopting diverse personas.
    }
    \label{fig:bug_issue_workflow}
\end{figure*}

\section{Realism-Oriented Synthesis: Bug Agent and Issue Agent}
\label{sec:bug_agent}

While high-throughput synthesis (Section~\ref{sec:swesmith}) is necessary for scale, it is insufficient for realism. Entity-local mutations tend to produce failures with a \textbf{local footprint}: the injected edit is in close proximity to the failing assertion, and the task description often inherits implementation details that leak the root cause. Real-world incidents exhibit distinct characteristics: regressions frequently stem from cross-module interactions, configuration-sensitive behavior, or violated assumptions at API boundaries, while issue reports typically emphasize \textbf{observable symptoms} and reproduction evidence rather than causal explanations. To bridge this realism gap (Figure~\ref{fig:bug_issue_workflow}), SWE-Hub introduces a dedicated \textbf{Realism Layer} comprising two specialized agents: the \textbf{Bug Agent}, which synthesizes \textbf{system-level regressions}, and the \textbf{Issue Agent}, which converts verified regressions into user-like issue reports under strict anti-leakage constraints.

\paragraph{Section contributions.}
This section makes two primary contributions: (i) the \textbf{Bug Agent}, a regression synthesis engine that targets system-level, non-local failures via contract-violating edits to mimic real-world debugging complexity; and (ii) the \textbf{Issue Agent}, a symptom-centric reporting module that enforces strict symptom-cause separation to generate leakage-resistant, user-like problem statements.

\subsection{Bug Agent: System-Level Regression Synthesis}
\label{sec:bug_agent_bug}

\paragraph{Operational definition.}
We define a \textbf{system-level regression} as a verified behavioral failure where the \textbf{cause, symptom, or both are non-local}. The injected change affects an interface, invariant, or cross-module contract such that failing tests manifest in components that are not co-located with the edit—often separated by file boundaries, module calls, or architectural layers. Unlike local mutation strategies, Bug Agent explicitly targets such regressions to maximize diagnostic depth.

\paragraph{Design principle: contract violations with ripple effects.}
Bug Agent is guided by a ``ripple effect'' principle: realistic regressions often originate from changes that appear \textbf{plausible from an engineering standpoint} (e.g., refactors, performance optimizations, or API cleanups) but subtly violate an \textbf{implicit contract} relied upon by downstream components. This creates failures that manifest remotely, thereby increasing diagnostic difficulty and better mimicking real-world debugging scenarios. Concretely, Bug Agent prioritizes edits that (i) preserve local syntactic validity and compilation, (ii) resemble coherent block-level modifications, and (iii) are likely to perturb cross-component behavior (e.g., altering boundary checks, default values, ordering guarantees, serialization formats, or configuration-sensitive branches).

\subsubsection{Workflow: Context--Edit--Verify Loop}
\label{sec:bug_agent_workflow}

Figure~\ref{fig:bug_issue_workflow} illustrates the end-to-end pipeline of this layer. Bug Agent follows a structured reasoning-verification loop grounded in execution:

\begin{enumerate}
    \item \textbf{Context audit and target selection.} The agent identifies candidate hotspots that are (a) exercised by the test suite and (b) structurally connected to other components (via callers, callees, shared data structures, or configuration paths). It then retrieves extended context, including relevant call sites and associated tests.
    \item \textbf{Motivation sampling (plausible rationale).} To steer edits toward realistic-looking patches rather than arbitrary corruption, the agent samples a refactoring ``motivation'' (e.g., performance optimization, code simplification, syntax modernization, or style consistency).
    \item \textbf{Block-level rewrite and injection.} The agent performs coherent block-level changes (targeting loops, conditionals, boundary handling, or API glue) to introduce a subtle semantic fault while maintaining plausible code structure.
    \item \textbf{Execution-based verification.} The injected patch is accepted only if it induces a pass-to-fail signal under the repository verifier (Section~\ref{sec:swesmith_k8s}). Verification yields structured traces ($T_{\text{P2F}}$, logs) used both for filtering and for downstream issue generation.
    \item \textbf{Artifact materialization.} The final output is a replayable regression instance: the bug injection patch, the pinned execution substrate reference, and verification metadata.
\end{enumerate}
This loop ensures that realism-oriented synthesis remains fully compatible with SWE-Hub’s reproducibility guarantees: all accepted regressions are executable and auditable.

\subsubsection{Regression Families}
\label{sec:bug_agent_families}

Bug Agent supports two complementary regression families that cover common real-world failure modes:

\paragraph{Single-change ripple regressions.}
A localized but contract-breaking edit triggers failures in downstream components (e.g., callers relying on a specific boundary condition, ordering, or default behavior). The symptom is separated from the edit site by call chains or module boundaries, modeling non-local debugging scenarios.

\paragraph{Co-modification (multi-module) regressions.}
Coordinated changes are applied across coupled modules (e.g., partial migrations, API reshaping, or cross-file refactors). Individually reasonable edits interact negatively to produce a regression, reflecting failures introduced during large-scale maintenance and architectural cleanup.

These families complement the local mutation synthesis of SWE-Scale: they increase both the causal depth and the surface diversity of failures without abandoning execution-grounded verification.

\subsection{Issue Agent: Symptom-Centric Reporting with Anti-Leakage}
\label{sec:issue_agent}

A realistic SWE task requires not only a buggy state but also a natural problem statement that mirrors how failures are reported in the wild. \textbf{Issue Agent} translates a verified regression (comprising a bug patch and failing evidence) into an issue-style prompt that emphasizes \textbf{symptoms} and \textbf{reproduction steps} while rigorously avoiding causal hints.

\subsubsection{Persona and Evidence Diversity}
\label{sec:issue_persona}

Real issue reports exhibit wide variance in tone and evidence. \textbf{Issue Agent} samples from a set of report personas and evidence styles to mitigate template artifacts, including: concise user reports, maintainer-style reports with structured steps, and log-centric reports dominated by stack traces or error messages. This diversity improves the robustness of downstream agent training and better aligns with the heterogeneous distribution of real-world reports.

\subsubsection{Symptom--Cause Separation as Constrained Generation}
\label{sec:issue_antileak}

Issue Agent enforces \textbf{symptom--root cause separation} through explicit constraints. Generated reports must:
\begin{itemize}
    \item describe observable behavior changes (expected vs.\ observed).
    \item include reproduction steps and minimal evidence (logs, failing test names where appropriate).
    \item avoid repair-oriented cues, such as naming the modified function/file as the cause, proposing specific fixes, or revealing implementation-level causal explanations.
\end{itemize}
We implement this as a post-generation validation step (a lightweight anti-leakage filter) that rejects reports violating the constraint set.

\paragraph{Minimal reproducible examples.}
When feasible, Issue Agent additionally synthesizes a minimal reproducible example (MRE) that isolates the symptom while remaining self-contained (e.g., using in-memory inputs). MRE generation is grounded in the same execution substrate and validated through the verifier interface, enhancing both realism and task usability.

\subsection{Produced Artifacts}
\label{sec:realism_artifacts}

The Realism Layer synthesizes a complete \textbf{Data Package} that instantiates the task family $f = \text{Bug Agent}$ within the unified task schema (Section~\ref{sec:task_schema}).
The \textbf{Bug Agent} provides the core regression artifacts: the problem patch $\mathcal{P}_{\text{bug}}$ (the injected system-level regression), the validation criteria $\mathcal{V}$ (derived from the P2F signal), and implicitly defines the oracle $\mathcal{O}$ as the original repository baseline (the clean state before injection).
The \textbf{Issue Agent} contributes the natural-language problem statement $\mathcal{P}$, ensuring the prompt is symptom-centric and leakage-resistant.
Combined with the repository spec $\mathcal{R}$ and pinned environment $\mathcal{E}$, this forms a self-contained, high-fidelity task ready for evaluating agent capabilities in realistic debugging scenarios.

\subsection{Limitations: The Verification Coverage Paradox and Future Directions}
\label{sec:bug_agent_limit}

As illustrated in Figure~\ref{fig:bug_issue_workflow}, the Realism Layer inherits a fundamental limitation of execution-grounded datasets: verification relies on an existing test oracle.
This introduces survivorship bias—only regressions covered by available tests can be validated and retained—while real-world failures sometimes escape CI due to inadequate coverage.

We refer to this tension as the \textbf{Verification Coverage Paradox}: the most severe and subtle regressions may be precisely those that are hardest to certify under existing tests.
A promising direction is \textbf{reverse test generation}: when a plausibly injected regression does not trigger any failing tests, the system generates a targeted reproduction test that fails on the buggy version but passes on the clean baseline.
This closes the loop by producing both a certified bug and a regression test artifact, improving coverage and task fidelity.

\begin{figure}[t]
    \centering

    \includegraphics[width=\linewidth]{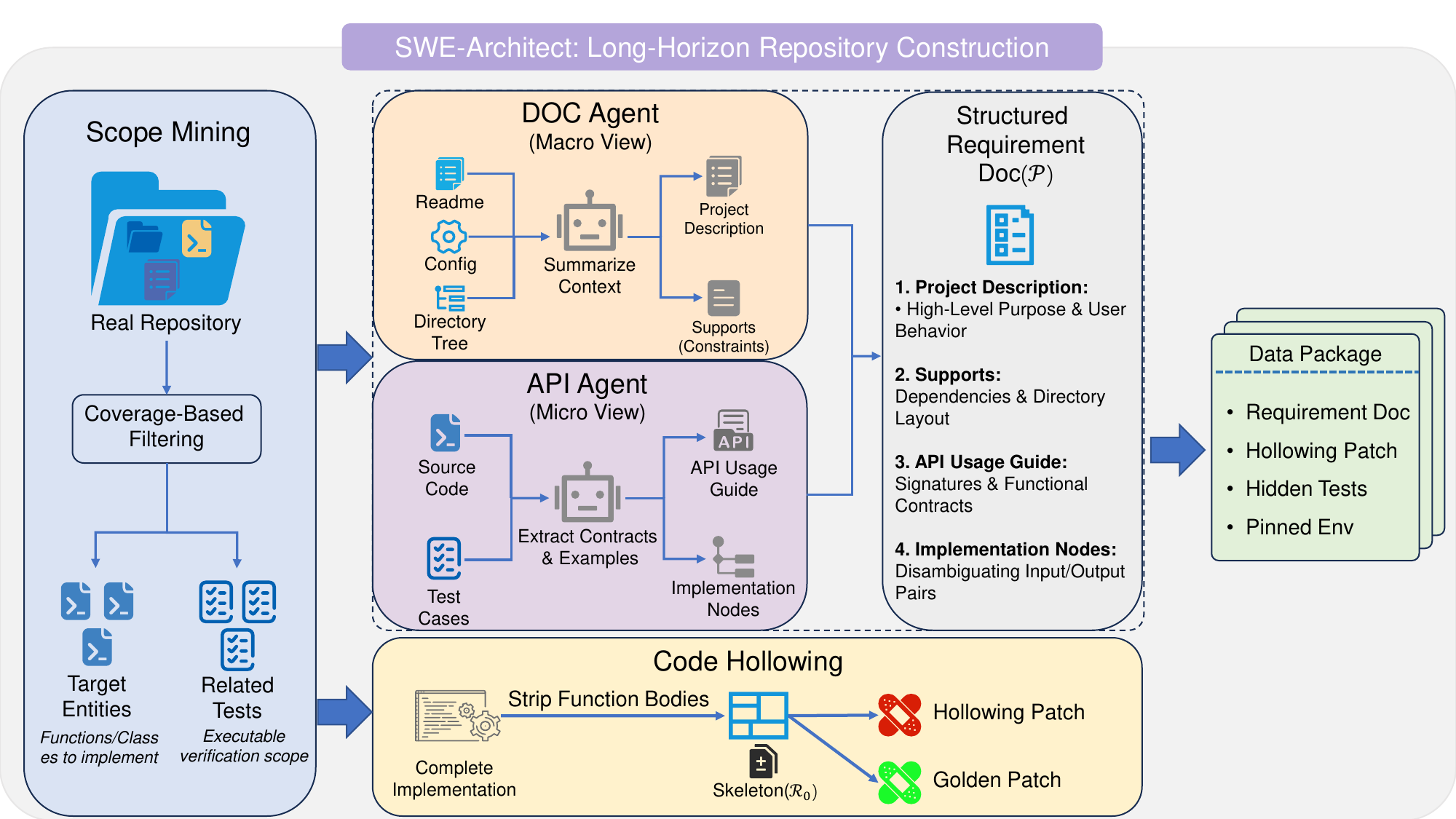} 
    \caption{\textbf{SWE-Architect pipeline.} (1) Mine a cohesive task scope from a real repository using coverage structure. (2) Create an initial repository state via code hollowing, and retain a golden reference patch. (3) Generate a structured requirement document via a two-agent reverse-engineering process (DOC Agent + API Agent). (4) Package the task with the pinned execution substrate and hidden tests in the unified task schema.}
    \label{fig:nl2repo_overview}
\end{figure}

\section{SWE-Architect: Long-Horizon Repository Construction}
\label{sec:nl2repo}

While repair-centric benchmarks provide strong signals for localized debugging, they systematically under-evaluate an agent’s capacity to \textbf{create}: translating high-level requirements into coherent, repository-scale software. In many existing settings, edit scopes are heavily constrained (e.g., single-file patches, fixed signatures), allowing agents to succeed without making architectural decisions, managing dependencies, or maintaining global invariants. \textbf{SWE-Architect} expands the task space from repair to \textbf{repository construction}. Each task presents a structured requirement document and an incomplete repository state; the agent must produce a functional repository implementation that satisfies an executable test oracle. SWE-Architect is designed to impose \textbf{long-horizon pressure} while remaining objectively verifiable and scalable to diverse real-world repositories.

SWE-Architect explicitly targets four recurring challenges in complex development workflows:
\begin{enumerate}
    \item \textbf{Planning \& termination} (deciding what to build and defining completion criteria),
    \item \textbf{Global coherence} (ensuring cross-file consistency and interface alignment),
    \item \textbf{Dependency management} (handling libraries, build/test tooling, and runtime constraints), and
    \item \textbf{Long-range reasoning} (linking high-level requirements to implementation details across architectural modules).
\end{enumerate}

\paragraph{Section contributions.}
This section makes three primary contributions: (i) the \textbf{SWE-Architect} framework, which extends the SWE-Hub's scope from repair to repository-scale construction, targeting long-horizon competencies like planning and dependency management; (ii) the \textbf{task mining pipeline}, utilizing code hollowing and coverage-based scope mining to generate executable, skeleton-based tasks from real repositories; and (iii) the \textbf{agentic document generation process}, employing DOC and API agents to reverse-engineer structured requirement documents that balance specificity with implementation leakage resistance.

\subsection{Task Definition and Execution Oracle}
\label{sec:nl2repo_oracle}

A SWE-Architect instance provides a structured natural-language specification $\mathcal{P}$ and an initial repository state $\mathcal{R}_0$ (typically a realistic skeleton with missing implementations). An agent produces a completed repository $\mathcal{R}' = \pi(\mathcal{P}, \mathcal{R}_0)$. Evaluation is strictly \emph{execution-grounded} via a hidden test oracle:
\[
V(\mathcal{R}') = I\big[\text{RunTests}(\mathcal{R}', \mathcal{E}, \mathcal{T}_{\text{hidden}})=\texttt{PASS}\big],
\]
where $\mathcal{E}$ is the pinned execution substrate (container image + standardized verifier entrypoint provided by the Environment Layer), and $\mathcal{T}_{\text{hidden}}$ is a private test suite. This definition ensures that success reflects repository-level functional correctness under a real verification harness, rather than surface-form similarity to a reference implementation.

\subsection{The Structured Requirement Document}
\label{sec:nl2repo_spec}

A central design choice is to encapsulate the task specification into a single structured document that is (i) sufficiently detailed to guide implementation, yet (ii) abstract enough to prevent trivial copying and root-cause leakage. Each document comprises four sections:

\begin{itemize}
    \item \textbf{Project Description}: The purpose, goals, and user-facing behavior of the system.
    \item \textbf{System Constraints}: Non-functional requirements including allowed dependencies, directory layout conventions, and runtime assumptions.
    \item \textbf{API Usage Guide}: The functional contract—defining public interfaces, expected behaviors, semantic constraints, and cross-API invariants.
    \item \textbf{Implementation Examples}: Concrete input/output examples for key APIs to disambiguate natural language requirements and clarify edge cases.
\end{itemize}

The document is written from an external, black-box perspective: it specifies interfaces and behaviors but deliberately excludes implementation code or internal causal explanations. This design encourages genuine planning and system design rather than simple template completion.

\subsection{From Real Repositories to SWE-Architect Instances}
\label{sec:nl2repo_factory}

SWE-Architect is produced by a scalable task factory that converts real repositories into construction tasks while preserving executable evaluation guarantees. Figure~\ref{fig:nl2repo_overview} summarizes the pipeline.

\subsubsection{Mining Cohesive Scope via Coverage Structure}
\label{sec:nl2repo_scope}

Directly requiring agents to rebuild entire repositories is neither scalable nor well-controlled. SWE-Architect instead derives a \emph{cohesive task scope} from each repository by analyzing the coverage structure (e.g., a test-to-code coverage graph) to identify clusters of functionality that are internally cohesive and relatively isolated. The output is a \textbf{candidate module package}: a precise list of target entities (functions/classes with file paths) to be implemented, mapped to the corresponding evaluation tests. To maintain a clean boundary between implementation and verification, entities originating from test files are strictly excluded from the target set.

\subsubsection{Initial State and Golden Patch via Code Hollowing}
\label{sec:nl2repo_hollow}

To create a realistic construction setting, we generate the initial repository state $\mathcal{R}_0$ through \textbf{code hollowing}: we remove the bodies of target entities while strictly preserving the surrounding architecture (file hierarchy, imports, type/interface declarations, and non-target code). This yields a repository that retains the structural complexity of a real codebase, requiring the agent to implement missing functionality within an existing context rather than starting from an empty directory.

We also retain a \textbf{golden patch} (a \texttt{git diff} computed from the original repository) that restores all hollowed code. The golden patch serves two purposes: (i) it certifies solvability by construction (guaranteeing a correct solution exists), and (ii) it provides a precise scope definition used for quality control and debugging of the task generation process.

\subsection{Agentic Document Generation: DOC Agent and API Agent}
\label{sec:nl2repo_doc_agents}

Synthesizing a high-quality requirement document is itself a reverse-engineering task: it requires both high-level project understanding and low-level API accuracy. SWE-Architect therefore employs a collaborative two-agent design.

\paragraph{DOC Agent (macro-level specification).}
DOC Agent extracts repository context (e.g., README, configuration, package metadata) to produce a PRD-like narrative covering the project purpose, functional goals, and infrastructure constraints (dependencies, directory conventions). It also organizes target entities into coherent functional groups, forming the structural backbone of the requirement document.

\paragraph{API Agent (micro-level contracts and examples).}
API Agent focuses on the target entities. For each entity, it derives:
(i) a static interface contract (signatures, type annotations where available, exceptions/edge cases), and
(ii) behavioral examples mined from tests or usage sites, reformatted as specification-level input/output nodes.
These artifacts populate the \textbf{API Usage Guide} and \textbf{Implementation Examples} sections, ensuring technical rigor while keeping the document implementation-agnostic.

This division of labor mirrors the abstraction gap in software engineering: repository context is broad and unstructured, while API behavior is local but demands precise technical detail.

\subsection{Packaging into SWE-Hub}
\label{sec:nl2repo_packaging}

The final output of SWE-Architect is a complete \textbf{Data Package} that instantiates the task family $f = \text{SWE-Architect}$ within the unified task schema (Section~\ref{sec:task_schema}).
The package aggregates the structured requirement document $\mathcal{P}$, the initial state defined by the hollowing patch $\mathcal{P}_{\text{bug}}$ (which removes implementations), and the golden patch $\mathcal{O}$ (the reference solution).
It also binds the validation criteria $\mathcal{V}$—the hidden test suite $\mathcal{T}_{\text{hidden}}$—to the pinned execution environment $\mathcal{E}$.
This configuration ensures that the construction task is fully executable and allows for direct comparison with repair-oriented tasks under the SWE-Hub's unified evaluation protocol.

\section{Conclusion}
\label{sec:conclusion}

We introduced \textbf{SWE-Hub}, a unified \emph{data factory} for software engineering agents that produces tasks that are \textbf{executable by construction}.
The core abstraction is a shared \emph{execution substrate}—pinned container images plus a standardized verifier entrypoint—provisioned automatically by the \textbf{Environment Layer} (Env Agent + Test Agent) to ensure reproducibility across heterogeneous repositories.
Built upon this substrate, SWE-Hub orchestrates three specialized product lines:
(i) \textbf{SWE-Scale}, a stateless synthesis-and-verification engine that enables Kubernetes-scale parallel validation and polyglot bug injection via a shared parsing backbone and configuration-as-code;
(ii) the \textbf{Realism Layer} (Bug Agent + Issue Agent), which targets \textbf{system-level regressions} and generates user-like issue reports with explicit \textbf{symptom--root cause separation} to mitigate leakage; and
(iii) \textbf{SWE-Architect}, which extends the evaluation horizon to \textbf{repository construction}, requiring agents to translate structured natural-language specifications into code validated against executable (often hidden) test oracles.

By treating SWE data generation as a modular production system rather than a one-off benchmark, SWE-Hub enables the continuous creation of diverse task families—spanning from short-horizon repair to long-horizon construction—under a consistent verification contract.
We posit that this factory paradigm establishes a robust foundation for more reliable training, more representative evaluation, and faster iteration of agentic SWE methods.
Future directions include expanding to broader ecosystem coverage, incorporating richer regression modes (e.g., performance, security, concurrency), and closing the loop with automated test generation and adaptive curricula driven by agent feedback.

\bibliography{custom}

\begin{thebibliography}{26}
\providecommand{\natexlab}[1]{#1}

\bibitem[{Chen et~al.(2026)Chen, Zhang, Feng, Wang, Zhao, Cao, Yang, Chen, Li, Ma, Ge, Zhang, Cui, Liu, Zhou, Sun, Lin, and Hui}]{swe-universe}
Mouxiang Chen, Lei Zhang, Yunlong Feng, Xuwu Wang, Wenting Zhao, Ruisheng Cao, Jiaxi Yang, Jiawei Chen, Mingze Li, Zeyao Ma, Hao Ge, Zongmeng Zhang, Zeyu Cui, Dayiheng Liu, Jingren Zhou, Jianling Sun, Junyang Lin, and Binyuan Hui. 2026.
\newblock \href {https://arxiv.org/abs/2602.02361} {Swe-universe: Scale real-world verifiable environments to millions}.
\newblock \emph{Preprint}, arXiv:2602.02361.

\bibitem[{Deng et~al.(2025)Deng, Da, Pan, He, Ide, Garg, Lauffer, Park, Pasari, Rane, Sampath, Krishnan, Kundurthy, Hendryx, Wang, Bharadwaj, Holm, Aluri, Zhang, Jacobson, Liu, and Kenstler}]{swebenchpro}
Xiang Deng, Jeff Da, Edwin Pan, Yannis~Yiming He, Charles Ide, Kanak Garg, Niklas Lauffer, Andrew Park, Nitin Pasari, Chetan Rane, Karmini Sampath, Maya Krishnan, Srivatsa Kundurthy, Sean Hendryx, Zifan Wang, Vijay Bharadwaj, Jeff Holm, Raja Aluri, Chen Bo~Calvin Zhang, and 3 others. 2025.
\newblock \href {https://arxiv.org/abs/2509.16941} {Swe-bench pro: Can ai agents solve long-horizon software engineering tasks?}
\newblock \emph{Preprint}, arXiv:2509.16941.

\bibitem[{Ding et~al.(2026)Ding, Long, Pu, Zhou, Gao, Gao, He, Hou, Hu, Li, Shi, Wang, Zan, Zhang, Zhang, Chen, Cheng, Deng, Gu, Hua, Lin, Liu, Li, Pan, Peng, Qin, Shan, Tan, Xie, Wang, Yuan, Zhang, Zhao, Zhao, Zhu, Zhu, Zou, Ding, Jiao, Liu, Liu, Liu, Tao, Yang, Yang, Zhang, Chen, Huang, and Zhang}]{nl2repobench}
Jingzhe Ding, Shengda Long, Changxin Pu, Huan Zhou, Hongwan Gao, Xiang Gao, Chao He, Yue Hou, Fei Hu, Zhaojian Li, Weiran Shi, Zaiyuan Wang, Daoguang Zan, Chenchen Zhang, Xiaoxu Zhang, Qizhi Chen, Xianfu Cheng, Bo~Deng, Qingshui Gu, and 30 others. 2026.
\newblock \href {https://arxiv.org/abs/2512.12730} {Nl2repo-bench: Towards long-horizon repository generation evaluation of coding agents}.
\newblock \emph{Preprint}, arXiv:2512.12730.

\bibitem[{Fu et~al.(2025)Fu, Liu, Shang, Ma, Yang, Liu, and Bian}]{multidockereval}
Kelin Fu, Tianyu Liu, Zeyu Shang, Yingwei Ma, Jian Yang, Jiaheng Liu, and Kaigui Bian. 2025.
\newblock \href {https://arxiv.org/abs/2512.06915} {Multi-docker-eval: A `shovel of the gold rush' benchmark on automatic environment building for software engineering}.
\newblock \emph{Preprint}, arXiv:2512.06915.

\bibitem[{He et~al.(2025)He, Liu, Du, Yan, Fan, Huang, Yuan, and Ma}]{swe-perf}
Xinyi He, Qian Liu, Mingzhe Du, Lin Yan, Zhijie Fan, Yiming Huang, Zejian Yuan, and Zejun Ma. 2025.
\newblock \href {https://arxiv.org/abs/2507.12415} {Swe-perf: Can language models optimize code performance on real-world repositories?}
\newblock \emph{Preprint}, arXiv:2507.12415.

\bibitem[{Jain et~al.(2025)Jain, Singh, Shetty, Zheng, Sen, and Stoica}]{r2egym}
Naman Jain, Jaskirat Singh, Manish Shetty, Liang Zheng, Koushik Sen, and Ion Stoica. 2025.
\newblock \href {https://arxiv.org/abs/2504.07164} {R2e-gym: Procedural environments and hybrid verifiers for scaling open-weights swe agents}.
\newblock \emph{Preprint}, arXiv:2504.07164.

\bibitem[{Jimenez et~al.(2024)Jimenez, Yang, Wettig, Yao, Pei, Press, and Narasimhan}]{swebench}
Carlos~E. Jimenez, John Yang, Alexander Wettig, Shunyu Yao, Kexin Pei, Ofir Press, and Karthik Narasimhan. 2024.
\newblock \href {https://arxiv.org/abs/2310.06770} {Swe-bench: Can language models resolve real-world github issues?}
\newblock \emph{Preprint}, arXiv:2310.06770.

\bibitem[{Kubernetes(2014)}]{kubernetes}
Kubernetes. 2014.
\newblock \href {https://kubernetes.io/} {Kubernetes: Production-grade container scheduling and management}.
\newblock Accessed 2026-02-05.

\bibitem[{{OpenAI}(2024)}]{swebenchverified}
{OpenAI}. 2024.
\newblock \href {https://openai.com/index/introducing-swe-bench-verified/} {Introducing swe-bench verified}.
\newblock Accessed 2026-02-05.

\bibitem[{Pan et~al.(2025)Pan, Wang, Neubig, Jaitly, Ji, Suhr, and Zhang}]{swegym}
Jiayi Pan, Xingyao Wang, Graham Neubig, Navdeep Jaitly, Heng Ji, Alane Suhr, and Yizhe Zhang. 2025.
\newblock \href {https://arxiv.org/abs/2412.21139} {Training software engineering agents and verifiers with swe-gym}.
\newblock \emph{Preprint}, arXiv:2412.21139.

\bibitem[{Pham et~al.(2025)Pham, Phan, Phan, Chi, Nguyen, and Bui}]{swesynth}
Minh V.~T. Pham, Huy~N. Phan, Hoang~N. Phan, Cuong~Le Chi, Tien~N. Nguyen, and Nghi D.~Q. Bui. 2025.
\newblock \href {https://arxiv.org/abs/2504.14757} {Swe-synth: Synthesizing verifiable bug-fix data to enable large language models in resolving real-world bugs}.
\newblock \emph{Preprint}, arXiv:2504.14757.

\bibitem[{{Qwen Team}()}]{qwen3codernext}
{Qwen Team}.
\newblock \href {https://github.com/QwenLM/Qwen3-Coder/blob/main/qwen3_coder_next_tech_report.pdf} {Qwen3-coder-next technical report}.
\newblock Technical report.
\newblock Accessed: 2026-02-03.

\bibitem[{Shum et~al.(2025)Shum, Hui, Chen, Zhang, W., Yang, Huang, Lin, and He}]{swe-rm}
KaShun Shum, Binyuan Hui, Jiawei Chen, Lei Zhang, X.~W., Jiaxi Yang, Yuzhen Huang, Junyang Lin, and Junxian He. 2025.
\newblock \href {https://arxiv.org/abs/2512.21919} {Swe-rm: Execution-free feedback for software engineering agents}.
\newblock \emph{Preprint}, arXiv:2512.21919.

\bibitem[{Song et~al.(2025)Song, Ramaneti, Sheikh, Chen, Gou, Xie, Xu, Zhang, Gandhi, Yang, Liu, Ou, Yuan, Xu, Zhou, Wang, Yue, Yu, Sun, Su, and Neubig}]{agentdataprotocol}
Yueqi Song, Ketan Ramaneti, Zaid Sheikh, Ziru Chen, Boyu Gou, Tianbao Xie, Yiheng Xu, Danyang Zhang, Apurva Gandhi, Fan Yang, Joseph Liu, Tianyue Ou, Zhihao Yuan, Frank Xu, Shuyan Zhou, Xingyao Wang, Xiang Yue, Tao Yu, Huan Sun, and 2 others. 2025.
\newblock \href {https://arxiv.org/abs/2510.24702} {Agent data protocol: Unifying datasets for diverse, effective fine-tuning of llm agents}.
\newblock \emph{Preprint}, arXiv:2510.24702.

\bibitem[{Tao et~al.(2026)Tao, Chen, Jiang, Kou, Wang, Wang, Li, Yang, Du, Dai, Mao, Wang, Shang, and Bai}]{swelego}
Chaofan Tao, Jierun Chen, Yuxin Jiang, Kaiqi Kou, Shaowei Wang, Ruoyu Wang, Xiaohui Li, Sidi Yang, Yiming Du, Jianbo Dai, Zhiming Mao, Xinyu Wang, Lifeng Shang, and Haoli Bai. 2026.
\newblock \href {https://arxiv.org/abs/2601.01426} {Swe-lego: Pushing the limits of supervised fine-tuning for software issue resolving}.
\newblock \emph{Preprint}, arXiv:2601.01426.

\bibitem[{Thai et~al.(2026)Thai, Le, Manh, Nhat, and Bui}]{sweevo}
Minh V.~T. Thai, Tue Le, Dung~Nguyen Manh, Huy~Phan Nhat, and Nghi D.~Q. Bui. 2026.
\newblock \href {https://arxiv.org/abs/2512.18470} {Swe-evo: Benchmarking coding agents in long-horizon software evolution scenarios}.
\newblock \emph{Preprint}, arXiv:2512.18470.

\bibitem[{Tree-sitter(2015)}]{tree-sitter}
Tree-sitter. 2015.
\newblock \href {https://github.com/tree-sitter/tree-sitter} {Tree-sitter: An incremental parsing system for programming tools}.
\newblock Accessed 2026-02-05.

\bibitem[{Wang et~al.(2025{\natexlab{a}})Wang, Zan, Xin, Liu, Wu, and Shen}]{swe-mirror}
Junhao Wang, Daoguang Zan, Shulin Xin, Siyao Liu, Yurong Wu, and Kai Shen. 2025{\natexlab{a}}.
\newblock \href {https://arxiv.org/abs/2509.08724} {Swe-mirror: Scaling issue-resolving datasets by mirroring issues across repositories}.
\newblock \emph{Preprint}, arXiv:2509.08724.

\bibitem[{Wang et~al.(2025{\natexlab{b}})Wang, Ramalho, Celestino, Pham, Liu, Sinha, Portillo, Osunwa, and Maduekwe}]{swebenchpp}
Lilin Wang, Lucas Ramalho, Alan Celestino, Phuc~Anthony Pham, Yu~Liu, Umang~Kumar Sinha, Andres Portillo, Onassis Osunwa, and Gabriel Maduekwe. 2025{\natexlab{b}}.
\newblock \href {https://arxiv.org/abs/2512.17419} {Swe-bench++: A framework for the scalable generation of software engineering benchmarks from open-source repositories}.
\newblock \emph{Preprint}, arXiv:2512.17419.

\bibitem[{Wei et~al.(2025)Wei, Sun, McMilin, Gehring, Zhang, Synnaeve, Fried, Zhang, and Wang}]{selfplayswerl}
Yuxiang Wei, Zhiqing Sun, Emily McMilin, Jonas Gehring, David Zhang, Gabriel Synnaeve, Daniel Fried, Lingming Zhang, and Sida Wang. 2025.
\newblock \href {https://arxiv.org/abs/2512.18552} {Toward training superintelligent software agents through self-play swe-rl}.
\newblock \emph{Preprint}, arXiv:2512.18552.

\bibitem[{Xu et~al.(2026)Xu, Yang, Tse-Hsun, and Chen}]{swe-refactor}
Yisen Xu, Jinqiu Yang, Tse-Hsun, and Chen. 2026.
\newblock \href {https://arxiv.org/abs/2602.03712} {Swe-refactor: A repository-level benchmark for real-world llm-based code refactoring}.
\newblock \emph{Preprint}, arXiv:2602.03712.

\bibitem[{Yang et~al.(2025)Yang, Lieret, Jimenez, Wettig, Khandpur, Zhang, Hui, Press, Schmidt, and Yang}]{swesmith}
John Yang, Kilian Lieret, Carlos~E. Jimenez, Alexander Wettig, Kabir Khandpur, Yanzhe Zhang, Binyuan Hui, Ofir Press, Ludwig Schmidt, and Diyi Yang. 2025.
\newblock \href {https://arxiv.org/abs/2504.21798} {Swe-smith: Scaling data for software engineering agents}.
\newblock \emph{Preprint}, arXiv:2504.21798.

\bibitem[{Zan et~al.(2025)Zan, Huang, Liu, Chen, Zhang, Xin, Chen, Liu, Zhong, Li, Liu, Xiao, Chen, Zhang, Su, Liu, Long, Shen, and Xiang}]{multiswebenche}
Daoguang Zan, Zhirong Huang, Wei Liu, Hanwu Chen, Linhao Zhang, Shulin Xin, Lu~Chen, Qi~Liu, Xiaojian Zhong, Aoyan Li, Siyao Liu, Yongsheng Xiao, Liangqiang Chen, Yuyu Zhang, Jing Su, Tianyu Liu, Rui Long, Kai Shen, and Liang Xiang. 2025.
\newblock \href {https://arxiv.org/abs/2504.02605} {Multi-swe-bench: A multilingual benchmark for issue resolving}.
\newblock \emph{Preprint}, arXiv:2504.02605.

\bibitem[{Zeng et~al.(2025)Zeng, Li, Xiao, Li, Liu, Yan, Wei, He, Song, Liu, and Zhou}]{skyworkswe}
Liang Zeng, Yongcong Li, Yuzhen Xiao, Changshi Li, Chris~Yuhao Liu, Rui Yan, Tianwen Wei, Jujie He, Xuchen Song, Yang Liu, and Yahui Zhou. 2025.
\newblock \href {https://arxiv.org/abs/2506.19290} {Skywork-swe: Unveiling data scaling laws for software engineering in llms}.
\newblock \emph{Preprint}, arXiv:2506.19290.

\bibitem[{Zhan et~al.(2025)Zhan, Deng, Tang, Xiang, Wu, Li, Zhu, Xu, Huang, Feng, Wang, Yan, Chen, Liu, Peng, Gao, Huang, Zhang, Wang, Lin, Li, Wang, Zhan, Wu, Zhang, Yang, Chen, Zhang, Chen, and Yu}]{katcoder}
Zizheng Zhan, Ken Deng, Huaixi Tang, Wen Xiang, Kun Wu, Weihao Li, Wenqiang Zhu, Jingxuan Xu, Lecheng Huang, Zongxian Feng, Shaojie Wang, Shangpeng Yan, Xuxing Chen, Jiaheng Liu, Zhongyuan Peng, Zuchen Gao, Haoyang Huang, Xiaojiang Zhang, Jinghui Wang, and 11 others. 2025.
\newblock \href {https://arxiv.org/abs/2507.08297} {Kat-v1: Kwai-autothink technical report}.
\newblock \emph{Preprint}, arXiv:2507.08297.

\bibitem[{Zhang et~al.(2025)Zhang, Yang, Yang, Yang, Chen, Zhang, Cui, Hui, and Lin}]{swe-flow}
Lei Zhang, Jiaxi Yang, Min Yang, Jian Yang, Mouxiang Chen, Jiajun Zhang, Zeyu Cui, Binyuan Hui, and Junyang Lin. 2025.
\newblock \href {https://arxiv.org/abs/2506.09003} {Swe-flow: Synthesizing software engineering data in a test-driven manner}.
\newblock \emph{Preprint}, arXiv:2506.09003.

\end{thebibliography}

\section*{Core Contributors}

\begin{multicols}{2}
    \begin{itemize}[noitemsep, topsep=0pt, leftmargin=*] 
        \item Yucheng Zeng\textsuperscript{$\ddagger$}
        \item Shupeng Li\textsuperscript{$\ddagger$}
        \item Daxiang Dong\textsuperscript{$\ddagger$}
        \item Ruijie Xu
        \item Zimo Chen
        \item Liwei Zheng
        \item Yuxuan Li
        \item Zhe Zhou
        \item Haotian Zhao
        \item Lun Tian
        \item Heng Xiao
        \item Tianshu Zhu
        \item Longkun Hao
        \item Jianmin Wu\textsuperscript{$\dagger$}
    \end{itemize}
\end{multicols}

\vspace{1em} 

\noindent \textsuperscript{$\ddagger$} Project Leader. \\
\textsuperscript{$\dagger$} Project Sponsor.
\end{document}